\PassOptionsToPackage{numbers, compress}{natbib}
\documentclass{article}

\usepackage[preprint]{neurips_2026}
\usepackage[utf8]{inputenc} 
\usepackage[T1]{fontenc}    
\usepackage{hyperref}       
\usepackage{url}            
\usepackage{booktabs}       
\usepackage{amsfonts}       
\usepackage{nicefrac}       
\usepackage{microtype}      
\usepackage{xcolor}         
\usepackage{amsmath} 
\usepackage{tabularx} 
\usepackage{array} 
\usepackage{graphicx} 
\usepackage{float}
\hypersetup{
    unicode=true,
    pdftitle={SHERPA: Seam-aware Harmonized ERP Adaptation for Open-Domain 360° Panorama Generation},
    pdfauthor={Jungwoon Kang and Jaehun Kim and Yiwon Yu and Hyungyum Jang and Sanghoon Lee and Jongyoo Kim},
    pdfkeywords={panorama generation, 360 panorama, diffusion model, flow matching, equirectangular projection}
}
\title{SHERPA: Seam-aware Harmonized ERP Adaptation for Open-Domain 360$^\circ$ Panorama Generation}

\author{%
  Jungwoon Kang \\
  Yonsei University \\
  \texttt{jw.kang@yonsei.ac.kr} \\
  \And
  Jaehun Kim \\
  Yonsei University \\
  \texttt{dbice@yonsei.ac.kr} \\
  \And
  Yiwon Yu \\
  Yonsei University \\
  \texttt{yw.yu@yonsei.ac.kr} \\
  \AND
  Hyungyum Jang \\
  Yonsei University \\
  \texttt{hyungyumjang@yonsei.ac.kr} \\
  \And
  Sanghoon Lee \\
  Yonsei University \\
  \texttt{slee@yonsei.ac.kr} \\
  \And
  Jongyoo Kim\thanks{Corresponding author.} \\
  Yonsei University \\
  \texttt{jy.kim@yonsei.ac.kr}
}

\begin{document}

\maketitle

\begin{figure}[H]
\centering
\includegraphics[width=\textwidth]{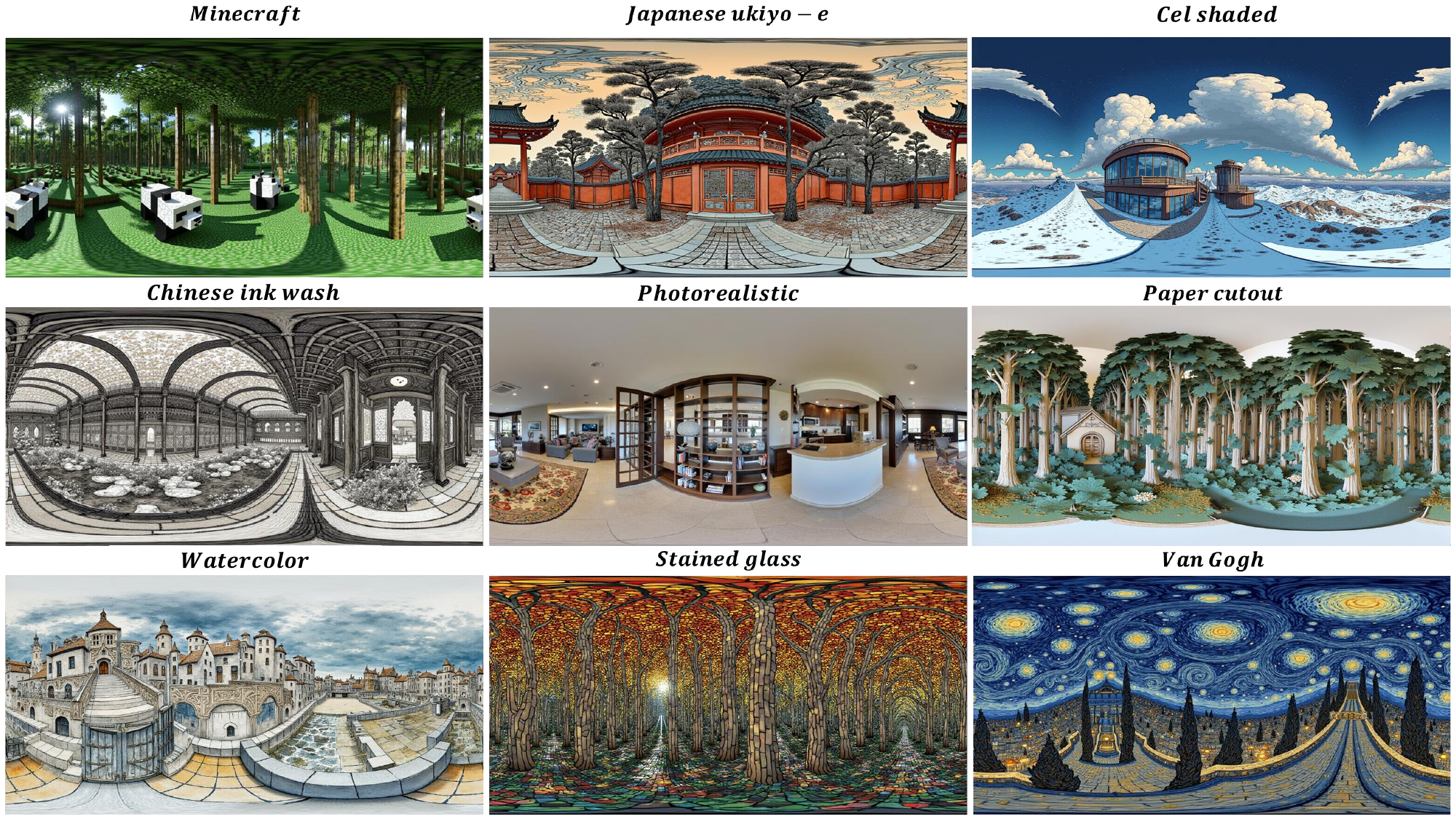}
\caption{\textbf{Open-domain panorama generation with SHERPA.}
SHERPA generates $360^\circ$ panoramas across photorealistic and stylized domains.}
\label{fig:teaser}
\end{figure}

\begin{abstract}
Panoramic imagery is increasingly used in world-generation, games, and simulation, where users may need not only photorealistic scenes but also stylized and non-photorealistic environments.
Large-scale text-to-image diffusion and flow models provide broad style and semantic priors for this goal, but planar image training misaligns them with the wrap-around topology and polar regions of $360^\circ$ panoramas represented in equirectangular projection (ERP).
We present SHERPA, a lightweight adaptation framework that combines frequency-selective Circular RoPE, Circular Latent Encoding/Decoding, image-side FFN adapters, and a Dual-Path Training Scheme.
Circular RoPE replaces only the seam-sensitive high-frequency horizontal RoPE band with integer-periodic harmonics while preserving the pretrained lower-frequency spectrum.
The Paired Panorama Path supervises geometry, while the Unpaired Style Path uses self-supervised yaw consistency for target-free stylized prompts.
As a result, SHERPA generates $360^\circ$ panoramas across both photorealistic panorama domains and open-domain stylized prompts.
\end{abstract}

\section{Introduction}

Open-domain panorama generation aims to create not only photorealistic $360^\circ$ scenes, but also stylized and user-specified immersive environments from text.
This capability is increasingly useful for world-model data generation, games, simulation, and spatial content creation, where diverse visual appearances are often as important as geometric coverage.
Large-scale text-to-image diffusion and flow models, especially DiT backbones such as SD3 and FLUX, provide strong prompt following, high fidelity, semantic understanding, and broad open-domain styles~\cite{peebles2023scalable,esser2024scaling,flux2024}, making them attractive foundations for this goal.

However, $360^\circ$ panoramas represented in equirectangular projection (ERP) are not ordinary planar images: they require horizontal wrap-around continuity and exhibit latitude-dependent polar distortion~\cite{wang2025survey,lin2025panorama}. Throughout this paper, \emph{panorama} denotes a $360^\circ$ ERP image unless otherwise stated.
Directly applying pretrained planar models often produces seam discontinuities, polar stretching, repeated structures, and global geometric inconsistency~\cite{feng2023diffusion360,ni2025makes,feng2025dit360}.
Existing panorama methods address these geometry issues with multi-view generation, cubemap or tangent-plane fusion, single-pass ERP training, circular padding, and spherical representations~\cite{kalischek2025cubediff,ccapuk2025tandit,zheng2025panorama,dreamcube,feng2023diffusion360,wu2024spherediffusion,park2025spherediff}.

The remaining challenge is that geometry adaptation and open-domain style preservation are coupled.
Paired panorama supervision is far narrower than web-scale planar image-text data and is dominated by photorealistic sources such as Matterport3D, SUN360, Laval HDR panoramas, and Poly Haven~\cite{chang2017matterport3d,xiao2012sun360,gardner2017learning,polyhaven}.
Adapting too strongly to these paired photorealistic panoramas can improve ERP structure, but it can also suppress the broad visual prior that makes pretrained text-to-image models useful for stylized prompts.
Thus, open-domain panorama adaptation should learn ERP topology and pole/global structure while retaining the pretrained model's open-domain visual prior.

To this end, we propose \textbf{SHERPA}, a \textbf{S}eam-aware \textbf{H}armonized \textbf{ERP} \textbf{A}daptation framework.
SHERPA combines denoising-time \emph{Circular RoPE}, \emph{Circular Latent Encoding/Decoding}~\cite{wu2026anything}, image-side FFN adapters, and a Dual-Path Training Scheme: paired panoramas provide supervised geometry losses, while unpaired style prompts provide yaw-equivariant consistency.
The adaptation is parameter-efficient: the reported adapter has 7,471,104 trainable parameters, about \(0.062\%\) of the 12B-parameter FLUX.1-dev backbone, and adds only 1.34\% inference overhead in our measured setting.

Our contributions are summarized as follows:
\begin{itemize}
    \item \textbf{Frequency-selective Circular RoPE for ERP topology.}
    We circularize the high-frequency width-axis RoPE band with integer-periodic harmonics while preserving the pretrained low-frequency band.

    \item \textbf{FFN adapter for ERP geometry adaptation.}
    We train lightweight image-side FFN residual adapters on frozen FLUX single-stream blocks for global panorama composition beyond seam continuity, optimizing only 7.47M parameters.

    \item \textbf{Dual-Path Training Scheme with self-supervised yaw consistency.}
    We combine a Paired Panorama Path with an Unpaired Style Path that encourages panoramic behavior without paired stylized panorama targets.
\end{itemize}

\section{Related Work}

\subsection{Diffusion Models and Flow Matching}

Early diffusion models generate images by reversing a Markovian noising process~\cite{ho2020denoising}. 
Recent foundation models such as FLUX and SD3 adopt rectified flow or flow matching, learning text-conditioned velocity fields with Transformer backbones over tokenized latents~\cite{liu2023flow,albergo2023building,dosovitskiyimage,peebles2023scalable,flux2024,esser2024scaling}. 
Given noise $x_0\sim\mathcal{N}(0,I)$ and data $x_1$, rectified flow uses
\begin{equation}
x_t=(1-t)x_0+t x_1,
\qquad
v^\ast=x_1-x_0 .
\label{eq:rf_background}
\end{equation}
Here, \(v^\ast\) denotes the ground-truth rectified-flow target velocity.
We build on this pretrained rectified-flow setting and focus on adapting spatial geometry to panoramas.

\subsection{Panorama Generation via Diffusion Models}

Early panorama diffusion methods extended image-plane field of view~\cite{bar2023multidiffusion,lee2023syncdiffusion}, but true panoramas require horizontal wrap-around continuity and latitude-dependent sampling geometry.

\paragraph{View decomposition and fusion.}
One strategy decomposes the panorama into perspective views, cubemap faces, or tangent planes and fuses them back to a panorama~\cite{kalischek2025cubediff,ccapuk2025tandit,zheng2025panorama,dreamcube,wang2023360}. 
This reuses strong 2D backbones and reduces local projection distortion, but requires multiple generations or view-fusion steps and can introduce cross-view inconsistency.

\paragraph{Direct panorama and spherical modeling.}
Direct panorama and spherical methods introduce circular padding, boundary blending, panorama objectives, manifold guidance, or spherical latents~\cite{feng2023diffusion360,wu2023panodiffusion,chen2022text2light,sun2025spherical,wu2024spherediffusion,park2025spherediff}; 360Anything traces seam artifacts to VAE zero-padding and uses circular latent encoding~\cite{wu2026anything}. 
Open-domain style prompts can still exhibit planar or wide-angle bias after adaptation.

\paragraph{Adapting pretrained foundation models.}
Recent methods increasingly adapt pretrained diffusion or DiT backbones, including DiT360, UniPano, PanFusion, HunyuanWorld, and DiffPano~\cite{feng2025dit360,ni2025makes,hulora,panfusion2024,team2025hunyuanworld,ye2024diffpano}. 
They show the value of pretrained backbones, but geometry adaptation can still entangle with style or semantic behavior.
SHERPA instead separates three roles: a non-trainable positional rewrite for the true horizontal topology, a lightweight image-token adapter for residual panorama-domain structure, and target-free style-prompt regularization without paired panorama supervision.

\section{Method}

\begin{figure}[t]
\centering
\includegraphics[width=\textwidth]{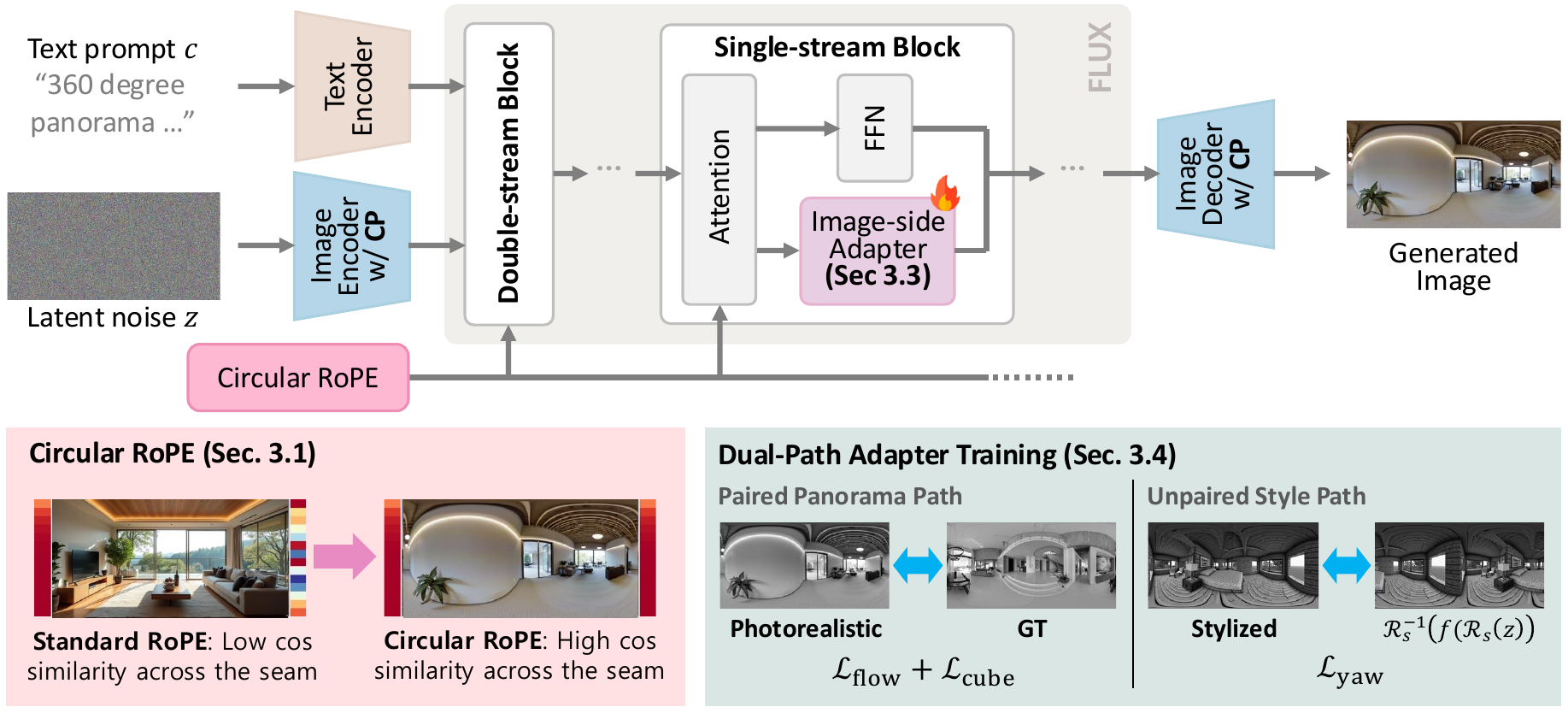}
\caption{\textbf{Overview of SHERPA.}
Frozen structural corrections are combined with trainable image-side FFN adapters.
Paired panoramas supervise flow and cubemap losses, while target-free style prompts provide yaw consistency. `CP' refers to circular padding.}
\label{fig:main_pipeline}
\end{figure}

Figure~\ref{fig:main_pipeline} gives an overview of SHERPA.
The generation path is shared: text and noise pass through the frozen FLUX transformer with Circular RoPE in attention and trainable image-side FFN adapters, followed by Circular Latent decoding.
Training separates supervision: paired panoramas provide flow and cubemap-projected velocity targets, while unpaired style prompts enforce yaw consistency between original and horizontally shifted latents.

\begin{figure}[t]
\centering
\includegraphics[width=0.98\textwidth]{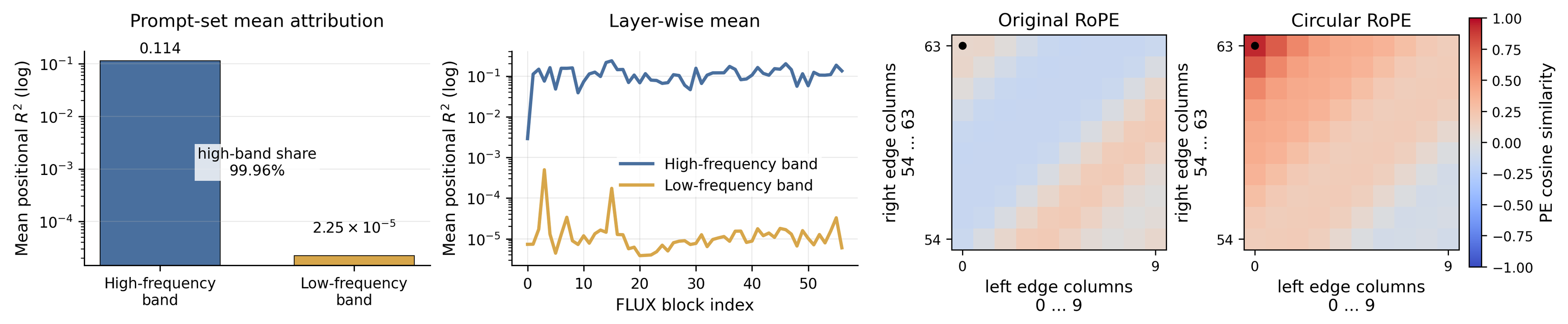}
\caption{\textbf{Analysis-guided Circular RoPE design.}
Left: prompt-set summary of head/layer/timestep-wise QK-logit regressions shows that measured width-axis positional response is concentrated in high-frequency RoPE bands.
Right: a seam-neighborhood PE similarity crop shows that replacing that band with integer-periodic harmonics makes the right and left panorama edges positionally adjacent.}
\label{fig:circular_rope_analysis_main}
\end{figure}

\subsection{Frequency-Selective Circular RoPE}
\label{sec:circular_rope}

Panoramas are horizontally periodic, but planar RoPE treats the first and last columns as distant grid positions.
Circular RoPE corrects this mismatch by replacing selected horizontal RoPE frequencies with integer-periodic harmonics over the panorama width.

For each selected width-axis RoPE pair with original frequency $\omega_j$, we replace it with the nearest integer-periodic harmonic over the packed longitude width $W$:
\begin{equation}
m_j =
\max\left(
1,
\mathrm{round}
\left(
\frac{\omega_j W}{2\pi}
\right)
\right),
\qquad
\tilde{\omega}_j =
\frac{2\pi m_j}{W}.
\label{eq:periodic_frequency}
\end{equation}
The resulting $\tilde{\omega}_j$ is the Circular RoPE frequency used for the selected width-axis pair.
This ensures horizontal phase closure for any horizontal coordinate $u$:
\begin{equation}\tilde{\omega}_j(u+W) = \tilde{\omega}_j u + 2\pi m_j . \label{eq:periodic_phase} \end{equation}

We do not rewrite the entire spectrum.
Given text prompts, we probe FLUX attention by regressing QK logits onto content and width-axis RoPE-band features, then measure how much of the positional component is explained by high- versus low-frequency bands.
Figure~\ref{fig:circular_rope_analysis_main} summarizes these head/layer/timestep-wise regressions over a small prompt set rather than showing a single attention heatmap.
The attribution assigns almost all measured positional response to the high-frequency width-axis RoPE band: the mean high-band positional $R^2$ is 0.1140, while the low-band attribution is only $2.25\times 10^{-5}$, and every measured head has higher high-band than low-band attribution.
Since this attribution is a diagnostic rather than a complete proof for all possible prompts, we use it to choose the candidate band and then validate the scope empirically with the sweep in Appendix~\ref{app:rope_scope_sweep}.
We directly replace only this high-frequency band with periodic harmonics, preserving lower frequencies while creating the desired left/right boundary compatibility.

\subsection{Circular Latent Decoding for Boundary Artifacts}
\label{sec:circular_vae}

Circular RoPE addresses denoising-time seam topology, but ordinary VAE convolutions can still introduce boundary artifacts.
Following 360Anything~\cite{wu2026anything}, we wrap horizontal boundary strips before VAE encoding/decoding and crop back to the original panorama width.
We evaluate this sequential structural effect in Sec.~\ref{sec:ablation}.

\subsection{Image-Side FFN Adapter for Panoramic Structure}
\label{sec:adapter}

\begin{figure}[t]
\centering
\includegraphics[width=0.92\textwidth]{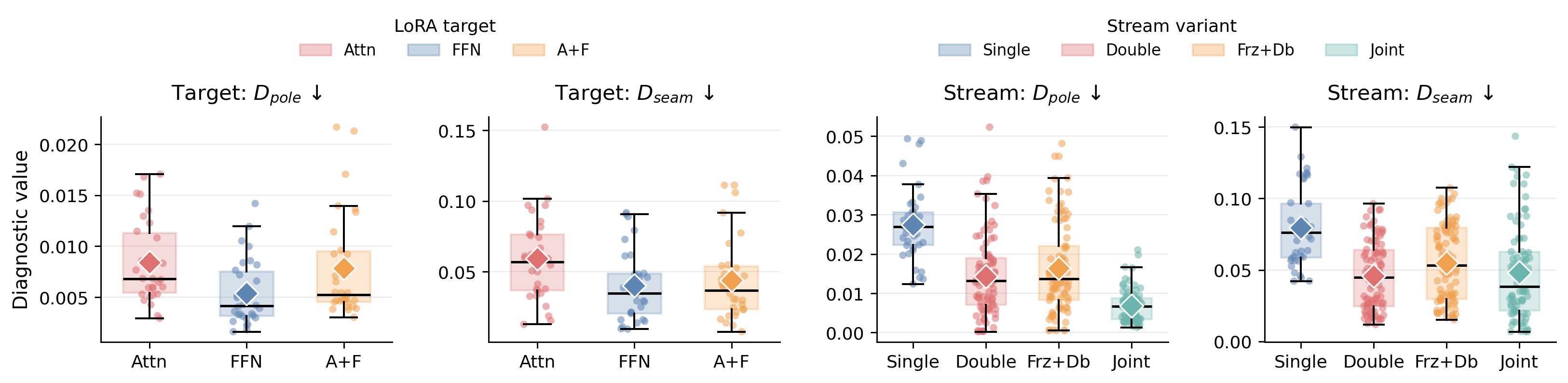}
\caption{\textbf{Adapter target diagnostic.}
Local seam and pole diagnostics guide where to place trainable capacity; points are generated samples and diamonds denote means.}
\label{fig:lora_target_diagnostic}
\end{figure}

Circular RoPE and Circular Latent Encoding/Decoding provide non-trainable seam biases, but they do not fully adapt FLUX to global panorama composition.
We therefore place trainable residual capacity in image-side FFN adapters of the FLUX single-stream blocks while keeping the backbone frozen.
This keeps attention modification structural: Circular RoPE changes token compatibility across the seam, while the FFN path learns residual image-token transformations after attention.
This choice is motivated by local target-placement diagnostics.
We compare attention projections, FFN projections, and their union using
$D_{\mathrm{seam}}$, the mean absolute RGB difference between left/right boundary strips, and
$D_{\mathrm{pole}}$, the average local image-gradient magnitude in top and bottom polar bands.
These diagnostics identify seam and polar artifacts but are not standalone perceptual quality metrics.
As shown in Fig.~\ref{fig:lora_target_diagnostic}, FFN-targeted variants give a favorable seam/pole diagnostic trade-off, and single-stream image-side adapters avoid the instability observed in double-stream variants.
We therefore reserve attention for the fixed Circular RoPE topology change and place learned panorama-domain capacity in the FFN path.

Let $h_l^{I}$ denote the image-token hidden states produced by the $l$-th single-stream block of the frozen transformer.
We attach an adapter to the block output:
\begin{equation}
\tilde{h}_{l}^{I}
=
h_l^{I}
+
A_{l,\mathrm{up}}
\sigma
\left(
A_{l,\mathrm{down}} h_l^{I}
\right),
\label{eq:image_adapter}
\end{equation}
where
$A_{l,\mathrm{down}}\in\mathbb{R}^{r\times d}$,
$A_{l,\mathrm{up}}\in\mathbb{R}^{d\times r}$,
$r \ll d$,
and $\sigma$ is a nonlinear activation.
The up projection is initialized to zero, so the adapted model initially behaves identically to the frozen base model.
Only the adapter parameters are optimized.

Thus, Circular RoPE defines seam-compatible positions, Circular Latent Encoding/Decoding protects the decode boundary, and the FFN adapter corrects image-token representations while leaving the backbone frozen.

\subsection{Dual-Path Adapter Training}
\label{sec:training_objective}

Training only on paired panorama targets improves photorealistic panoramas but can pull stylized prompts toward the supervised panorama domain.
We therefore use two training paths rather than separate network modules: paired panoramas use flow and cubemap-projected velocity targets, while target-free style prompts use self-supervised yaw consistency.
The \emph{Paired Panorama Path} learns velocity fields from targets, while the \emph{Unpaired Style Path} encourages horizontal panoramic behavior without paired stylized images.

\paragraph{Paired Panorama Path.}
Given a panorama training image $x_{\mathrm{pano}}$, we encode it as $z_{\mathrm{data}}=E(x_{\mathrm{pano}})$ and use Eq.~\eqref{eq:rf_background} with $x_0=\epsilon$ and $x_1=z_{\mathrm{data}}$.
For a panorama prompt $c_{\mathrm{pano}}$, the adapted model predicts
\begin{equation}
\hat{v}_{\psi}
=
f_{\theta,\psi}(z_t,t,c_{\mathrm{pano}}),
\end{equation}
where \(f_{\theta,\psi}\) denotes the adapted rectified-flow network, \(\theta\) denotes the frozen FLUX weights, and \(\psi\) denotes the adapter parameters.
Here, \(\hat{v}_{\psi}\) is the adapter-predicted velocity field and \(v^\ast=z_{\mathrm{data}}-\epsilon\) is the rectified-flow target velocity in latent space.
The main flow-matching term is
\begin{equation}
\mathcal{L}_{\mathrm{flow}}
=
\mathbb{E}
\left[
\left\|
\hat{v}_{\psi}-v^\ast
\right\|_2^2
\right].
\label{eq:adapter_flow_loss}
\end{equation}

We also compare predicted and target velocity fields after cubemap projection.
Let $C(\cdot)$ denote panorama-to-cubemap projection applied to the latent velocity map.
The cubemap-projected velocity loss is
\begin{equation}
\mathcal{L}_{\mathrm{cube}}
=
\mathbb{E}
\left[
\left\|
C(\hat{v}_{\psi})-C(v^\ast)
\right\|_2^2
\right].
\label{eq:cube_velocity_loss}
\end{equation}
This loss supervises denoising velocities in cubemap views rather than RGB reconstructions, emphasizing projection-aware consistency where panorama distortion is severe.
In practice, it also stabilizes adapter training by anchoring the predicted velocity field to image-like panorama targets under an additional projection.

\paragraph{Unpaired Style Path.}
Stylized prompts have no paired panorama training images, so we impose horizontal equivariance on random latents.
Let $z_{\mathrm{sty}}$ denote a random latent sampled for a style prompt $c_{\mathrm{sty}}$, and let $\mathcal{R}_{\Delta}(\cdot)$ denote the horizontal cyclic-roll operator by a random shift $\Delta$ along the panorama width, with inverse $\mathcal{R}_{\Delta}^{-1}=\mathcal{R}_{-\Delta}$.
We apply the same style prompt to the original and rolled latents and compare the predicted velocity fields after undoing the roll:
\begin{equation}
\mathcal{L}_{\mathrm{yaw}}
=
\mathbb{E}
\left[
\left\|
f_{\theta,\psi}(z_{\mathrm{sty}},t,c_{\mathrm{sty}})
-
\mathcal{R}_{\Delta}^{-1}\!\left(
f_{\theta,\psi}(\mathcal{R}_{\Delta}(z_{\mathrm{sty}}),t,c_{\mathrm{sty}})
\right)
\right\|_2^2
\right].
\label{eq:yaw_equiv_loss}
\end{equation}
Unlike supervised yaw losses on panorama samples, the Unpaired Style Path compares two adapted predictions because no stylized panorama target exists.
During this roll, the panorama coordinate ids are kept fixed, so the loss penalizes reliance on a privileged horizontal image center rather than merely testing coordinate-consistent shifting.
The term targets horizontal azimuthal uniformity rather than full spherical geometry; we analyze its effect in Sec.~\ref{sec:ablation}.

\textbf{Final objective.}
Combining the two paths, the adapter is trained with
\begin{equation}
\mathcal{L}
=
\mathcal{L}_{\mathrm{pano}}
+
\mathcal{L}_{\mathrm{style}},
\qquad
\mathcal{L}_{\mathrm{pano}}
=
\mathcal{L}_{\mathrm{flow}}
+
\lambda_{\mathrm{cube}}\mathcal{L}_{\mathrm{cube}},
\qquad
\mathcal{L}_{\mathrm{style}}
=
\lambda_{\mathrm{yaw}}\mathcal{L}_{\mathrm{yaw}} .
\label{eq:final_loss}
\end{equation}

\section{Experiments}

\begin{figure}[t]
\centering
\includegraphics[width=\textwidth]{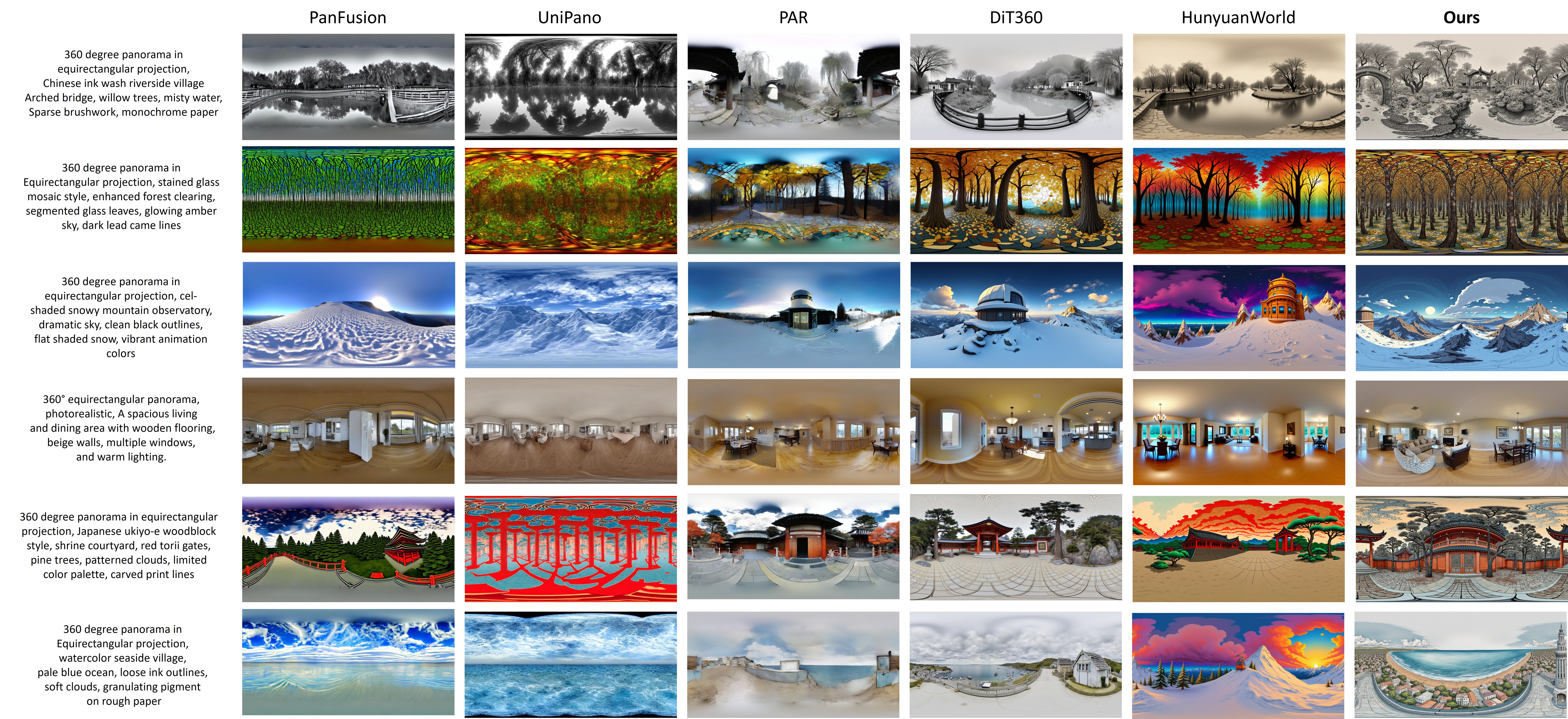}
\caption{\textbf{Qualitative comparison on open-domain panorama generation.}
SHERPA better preserves requested appearance and coherent panorama structure.}
\label{fig:qual_comparison}
\end{figure}

\subsection{Experimental Setup}
\label{sec:exp_setup}

\paragraph{Baselines.}
We compare SHERPA with representative panorama generation methods, including DiT360~\cite{feng2025dit360}, Diffusion360~\cite{feng2023diffusion360}, HunyuanWorld~\cite{team2025hunyuanworld}, UniPano~\cite{ni2025makes}, SMGD~\cite{sun2025spherical}, PanFusion~\cite{panfusion2024}, PanoWan~\cite{xia2026panowan}, PAR~\cite{wang2026conditional}, and Viewpoint~\cite{fang2025viewpoint}.
We use the same prompt list and fixed seeds whenever supported by the officially released pipelines; outputs are evaluated after conversion or resizing to \(1024{\times}512\) panoramas with the same preprocessing.

\paragraph{Metrics.}
We report FID~\cite{heusel2017gans}, Inception Score (IS)~\cite{salimans2016improved}, CLIPScore, and polar FID.
FID and IS use cubemap faces, CLIPScore uses the resized full panorama, and polar FID computes FID on the top and bottom cubemap faces.
Style-focused analyses additionally use qualitative comparison, a user study, and style-text CLIP.
For ablations, Seam denotes the mean absolute RGB difference between the left and right boundary strips, and Pole denotes the average image-gradient magnitude in the top and bottom polar bands.

\paragraph{Implementation.}
We build SHERPA on a pretrained FLUX rectified-flow DiT backbone~\cite{flux2024,esser2024scaling}.
Unless otherwise stated, Circular RoPE circularizes the high-frequency horizontal RoPE band, Circular Latent Encoding/Decoding is used for panorama latents, and FFN adapters are trained with the Dual-Path objective in Eq.~\ref{eq:final_loss}.
For paired panorama supervision, we use the polished Matterport3D version/split used by DiT360~\cite{feng2025dit360}.
Additional compute and asset-license details are reported in Appendix~\ref{app:compute_assets}.

\subsection{Main Comparison}
\label{sec:main_results}

We compare SHERPA with representative panorama baselines using automatic metrics on the shared Matterport test subset and qualitative/user-study evaluation on open-domain style prompts.
Figure~\ref{fig:qual_comparison} shows that SHERPA better balances requested appearance, panorama layout, and left-right continuity across both settings.

Table~\ref{tab:main_quant} reports multi-seed quantitative results on the shared Matterport test subset.
SHERPA obtains the best FID and polar FID while remaining competitive in IS and full-panorama CLIPScore.
HunyuanWorld has the highest CLIPScore but much worse polar FID, indicating that global text-image alignment alone can miss polar-region failures.
This matches our goal: adapting a strong pretrained image prior to panoramas while improving the distribution and polar structure of generated panoramas.

\begin{table}[t]
\centering
\scriptsize
\renewcommand{\arraystretch}{1.15}
\setlength{\tabcolsep}{3.4pt}
\begin{tabularx}{\textwidth}{@{}l *{4}{>{\centering\arraybackslash}X}@{}}
\toprule
\textbf{Method}
& \textbf{FID}$\downarrow$
& \textbf{IS}$\uparrow$
& \textbf{CLIP}$\uparrow$
& \textbf{Polar FID}$\downarrow$\\
\midrule
Diffusion360~\cite{feng2023diffusion360} & 56.45 $\pm$ 3.77 & 6.86 $\pm$ 0.36 & 28.78 $\pm$ 0.40 & 95.83 $\pm$ 4.07\\
DiT360~\cite{feng2025dit360} & 32.67 $\pm$ 1.01 & 7.28 $\pm$ 0.19 & \underline{30.96 $\pm$ 0.23} & \underline{79.16 $\pm$ 3.12}\\
PanFusion~\cite{panfusion2024} & 51.63 $\pm$ 0.44 & \underline{8.32 $\pm$ 0.05} & 29.12 $\pm$ 0.06 & 110.82 $\pm$ 2.19\\
SMGD~\cite{sun2025spherical} & 38.76 $\pm$ 1.49 & 5.21 $\pm$ 0.17 & 27.60 $\pm$ 0.30 & 98.63 $\pm$ 4.87\\
UniPano~\cite{ni2025makes} & 69.13 $\pm$ 0.67 & 7.84 $\pm$ 0.12 & 29.40 $\pm$ 0.09 & 154.75 $\pm$ 2.05\\
PanoWan~\cite{xia2026panowan} & 44.24 $\pm$ 0.34 & \textbf{8.56 $\pm$ 0.01} & 30.57 $\pm$ 0.08 & 94.12 $\pm$ 0.07\\
PAR~\cite{wang2026conditional} & \underline{32.73 $\pm$ 0.41} & 7.74 $\pm$ 0.07 & 30.73 $\pm$ 0.17 & 94.00 $\pm$ 0.66\\
Viewpoint~\cite{fang2025viewpoint} & 62.08 $\pm$ 0.42 & 7.53 $\pm$ 0.00 & 28.38 $\pm$ 0.02 & 128.87 $\pm$ 0.41\\
HunyuanWorld~\cite{team2025hunyuanworld} & 63.30 $\pm$ 1.96 & 6.29 $\pm$ 0.19 & \textbf{31.54 $\pm$ 0.18} & 124.49 $\pm$ 6.08\\
\midrule
\textbf{SHERPA (Ours)} & \textbf{27.80 $\pm$ 0.24} & 7.42 $\pm$ 0.15 & 30.75 $\pm$ 0.01 & \textbf{66.29 $\pm$ 0.23}\\
\bottomrule
\end{tabularx}
\caption{\textbf{Multi-seed quantitative comparison.}
Metrics are averaged across multiple generation seeds on the held-out prompt set.
FID/IS use cubemap faces, CLIP uses the full panorama, and polar FID uses the top and bottom cubemap faces.
Best values are bolded and second-best values are underlined.}
\label{tab:main_quant}
\end{table}
\textbf{Open-domain style generation.}
We evaluate style-conditioned prompts covering animation, painting, voxel-like scenes, and other non-photorealistic domains.
Because full-panorama CLIP can miss style-specific failures, we additionally compute CLIP similarity against style-only text phrases on the same 44 outputs per method.
SHERPA ranks second overall behind HunyuanWorld and above the other panorama baselines, showing that style-text alignment alone does not capture panorama structure or human preference.

\textbf{User study.}
We therefore pair automatic diagnostics with an anonymized user study on mixed photorealistic and stylized prompts.
The study was recruited through an open call, covered five prompt/style settings and 10 model outputs per setting, and asked raters to score 360-degree format consistency, reference-style consistency, and text-image consistency on a 1--5 Likert scale after reporting image-generation experience.
Self-reported rater experience was 69\% none/beginner, 27\% intermediate/advanced, and 4\% unspecified.
Fleiss' \(\kappa\) over all ratings is 0.091 (0.043/0.111/0.102 for panorama/style/text); paired respondent-prompt tests show SHERPA significantly exceeds DiT360 overall (mean difference \(0.35\), bootstrap 95\% CI \([0.26,0.44]\), \(p<10^{-13}\)).
The user study ranks SHERPA highest overall, supporting the qualitative comparison in Fig.~\ref{fig:qual_comparison}.

\begin{table}[t]
\centering
\scriptsize
\renewcommand{\arraystretch}{1.12}
\setlength{\tabcolsep}{2.2pt}
\begin{tabularx}{\textwidth}{@{}l *{5}{>{\centering\arraybackslash}X}@{}}
\toprule
\textbf{Method}
& \textbf{360 Match}$\uparrow$
& \textbf{Style Match}$\uparrow$
& \textbf{Text Match}$\uparrow$
& \textbf{Overall}$\uparrow$
& \textbf{Style-CLIP}$\uparrow$ \\
\midrule
Diffusion360~\cite{feng2023diffusion360} & 3.41 $\pm$ 0.07 & 2.57 $\pm$ 0.07 & 3.03 $\pm$ 0.07 & 3.00 $\pm$ 0.06 & 23.96 $\pm$ 0.85 \\
DiT360~\cite{feng2025dit360} & \underline{3.74 $\pm$ 0.06} & \underline{3.42 $\pm$ 0.07} & \underline{3.73 $\pm$ 0.06} & \underline{3.63 $\pm$ 0.05} & 23.51 $\pm$ 0.70 \\
PanFusion~\cite{panfusion2024} & 3.24 $\pm$ 0.07 & 3.02 $\pm$ 0.07 & 2.89 $\pm$ 0.06 & 3.05 $\pm$ 0.05 & 23.86 $\pm$ 0.60 \\
PanoWan~\cite{xia2026panowan} & 3.25 $\pm$ 0.07 & 2.81 $\pm$ 0.07 & 2.94 $\pm$ 0.07 & 3.00 $\pm$ 0.06 & 24.51 $\pm$ 0.69 \\
PAR~\cite{wang2026conditional} & 3.71 $\pm$ 0.06 & 3.09 $\pm$ 0.07 & 3.46 $\pm$ 0.06 & 3.42 $\pm$ 0.05 & 23.68 $\pm$ 0.70 \\
SMGD~\cite{sun2025spherical} & 3.48 $\pm$ 0.07 & 2.17 $\pm$ 0.08 & 2.14 $\pm$ 0.07 & 2.60 $\pm$ 0.06 & 17.88 $\pm$ 0.28 \\
UniPano~\cite{ni2025makes} & 2.82 $\pm$ 0.07 & 2.63 $\pm$ 0.07 & 2.50 $\pm$ 0.07 & 2.65 $\pm$ 0.06 & 24.06 $\pm$ 0.77 \\
Viewpoint~\cite{fang2025viewpoint} & 2.93 $\pm$ 0.07 & 2.21 $\pm$ 0.06 & 2.27 $\pm$ 0.06 & 2.47 $\pm$ 0.06 & 23.60 $\pm$ 0.57 \\
HunyuanWorld~\cite{team2025hunyuanworld} & 3.31 $\pm$ 0.06 & 2.91 $\pm$ 0.07 & 3.31 $\pm$ 0.06 & 3.18 $\pm$ 0.05 & \textbf{26.32 $\pm$ 0.34} \\
\midrule
\textbf{SHERPA (Ours)} & \textbf{3.98 $\pm$ 0.06} & \textbf{3.97 $\pm$ 0.05} & \textbf{3.99 $\pm$ 0.05} & \textbf{3.98 $\pm$ 0.05} & \underline{25.95 $\pm$ 0.64} \\
\bottomrule
\end{tabularx}
\caption{\textbf{User study on anonymized panorama outputs.}
Seventy raters score panorama, style, text, and overall consistency on a 1--5 Likert scale.
Style-CLIP is computed on 44 matched outputs per method using style-only text phrases.
We report mean $\pm$ standard error; Fleiss' \(\kappa=0.091\) over all human-rated criteria.}
\label{tab:style_eval}
\end{table}

\begin{table}[t]
\centering
\scriptsize
\renewcommand{\arraystretch}{1.15}
\setlength{\tabcolsep}{3.2pt}

\begin{tabularx}{\textwidth}{@{}
>{\raggedright\arraybackslash}p{0.15\textwidth}
>{\raggedright\arraybackslash}X
>{\centering\arraybackslash}p{0.065\textwidth}
>{\centering\arraybackslash}p{0.065\textwidth}
>{\centering\arraybackslash}p{0.065\textwidth}
>{\centering\arraybackslash}p{0.095\textwidth}
>{\centering\arraybackslash}p{0.105\textwidth}
@{}}
\toprule
\textbf{Variant}
& \textbf{Loss}
& \textbf{Seam}$\downarrow$
& \textbf{Polar FID}$\downarrow$
& \textbf{FID}$\downarrow$
& \textbf{IS}$\uparrow$
& \textbf{CLIP}$\uparrow$ \\
\midrule

Base FLUX
& structural only
& 0.145 & n/a & 47.07 & 7.55 & \underline{26.26} \\

RoPE only
& structural only
& \underline{0.068} & n/a & \textbf{43.20} & 7.54 & 26.01 \\

Latent E/D only
& structural only
& 0.132 & n/a & 47.23 & \underline{7.57} & \textbf{26.28} \\

RoPE + Latent E/D
& structural only
& \textbf{0.053} & n/a & \underline{43.84} & \textbf{7.62} & 25.78 \\

\midrule
\multicolumn{7}{@{}l}{\emph{Adapter and loss ablation}}\\
\midrule

Flow only
& $\mathcal{L}_{\mathrm{flow}}$
& \underline{0.0553} & \underline{71.29} & \underline{30.42} & 7.37 $\pm$ 0.52 & \underline{32.81 $\pm$ 2.48} \\

Flow + yaw, no cube
& $\mathcal{L}_{\mathrm{flow}}+\lambda_{\mathrm{yaw}}\mathcal{L}_{\mathrm{yaw}}$
& 0.2172 & 129.33 & 107.87 & 5.39 $\pm$ 0.27 & 26.15 $\pm$ 7.80 \\

Cube only
& $\mathcal{L}_{\mathrm{flow}}+\lambda_{\mathrm{cube}}\mathcal{L}_{\mathrm{cube}}$
& \textbf{0.0706} & \textbf{68.79} & \textbf{28.92} & 7.37 $\pm$ 0.47 & 32.61 $\pm$ 2.49 \\

Cube0.5 + yaw50
& $\mathcal{L}_{\mathrm{flow}}+\lambda_{\mathrm{cube}}\mathcal{L}_{\mathrm{cube}}+\lambda_{\mathrm{yaw}}\mathcal{L}_{\mathrm{yaw}}$
& 0.0878 & 72.90 & 30.89 & \underline{7.52 $\pm$ 0.29} & \textbf{32.79 $\pm$ 2.40} \\

Cube1 + yaw60
& high $\lambda_{\mathrm{cube}},\lambda_{\mathrm{yaw}}$
& 0.1686 & 118.00 & 92.57 & 6.55 $\pm$ 0.34 & 28.13 $\pm$ 8.10 \\

\textbf{SHERPA}
& \textbf{Eq.~\ref{eq:final_loss}}
& 0.0959 & 71.64 & 30.61 & \textbf{7.73 $\pm$ 0.39} & 32.30 $\pm$ 2.54 \\

\bottomrule
\end{tabularx}
\vspace{0.5pt}
\caption{\textbf{Ablation of SHERPA components and losses.}
The upper block isolates structural components; the lower block varies adapter losses.
Seam is boundary MAE; polar FID is computed on the top and bottom cubemap faces.
\textbf{Cube only scores best on GT-based photorealistic metrics, whereas Fig.~\ref{fig:loss_ablation} shows weaker stylized prompt fidelity; SHERPA is therefore selected as the preferred geometry--style trade-off.}
Best and second-best values are marked within each block.}
\label{tab:component_ablation}
\end{table}

\subsection{Ablation Studies}
\label{sec:ablation}

\begin{figure}[t]
\centering
\includegraphics[width=\textwidth]{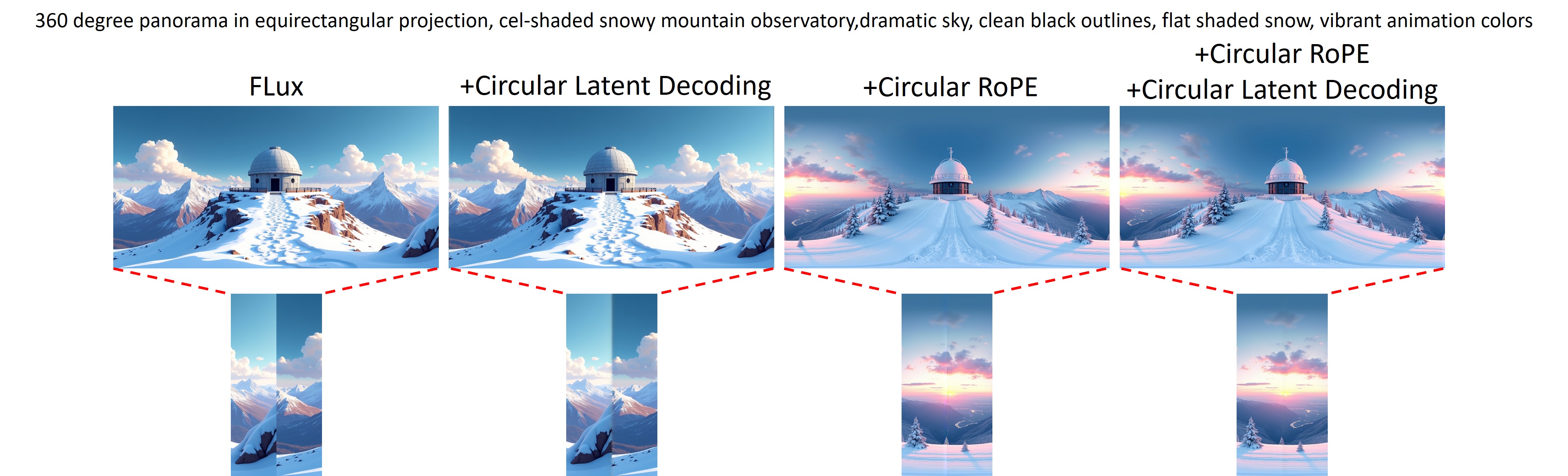}
\caption{\textbf{Structural component ablation.}
Visual comparison of the non-trainable panorama components used by SHERPA.}
\label{fig:parts_ablation}
\end{figure}

\begin{figure}[t]
\centering
\includegraphics[width=\textwidth]{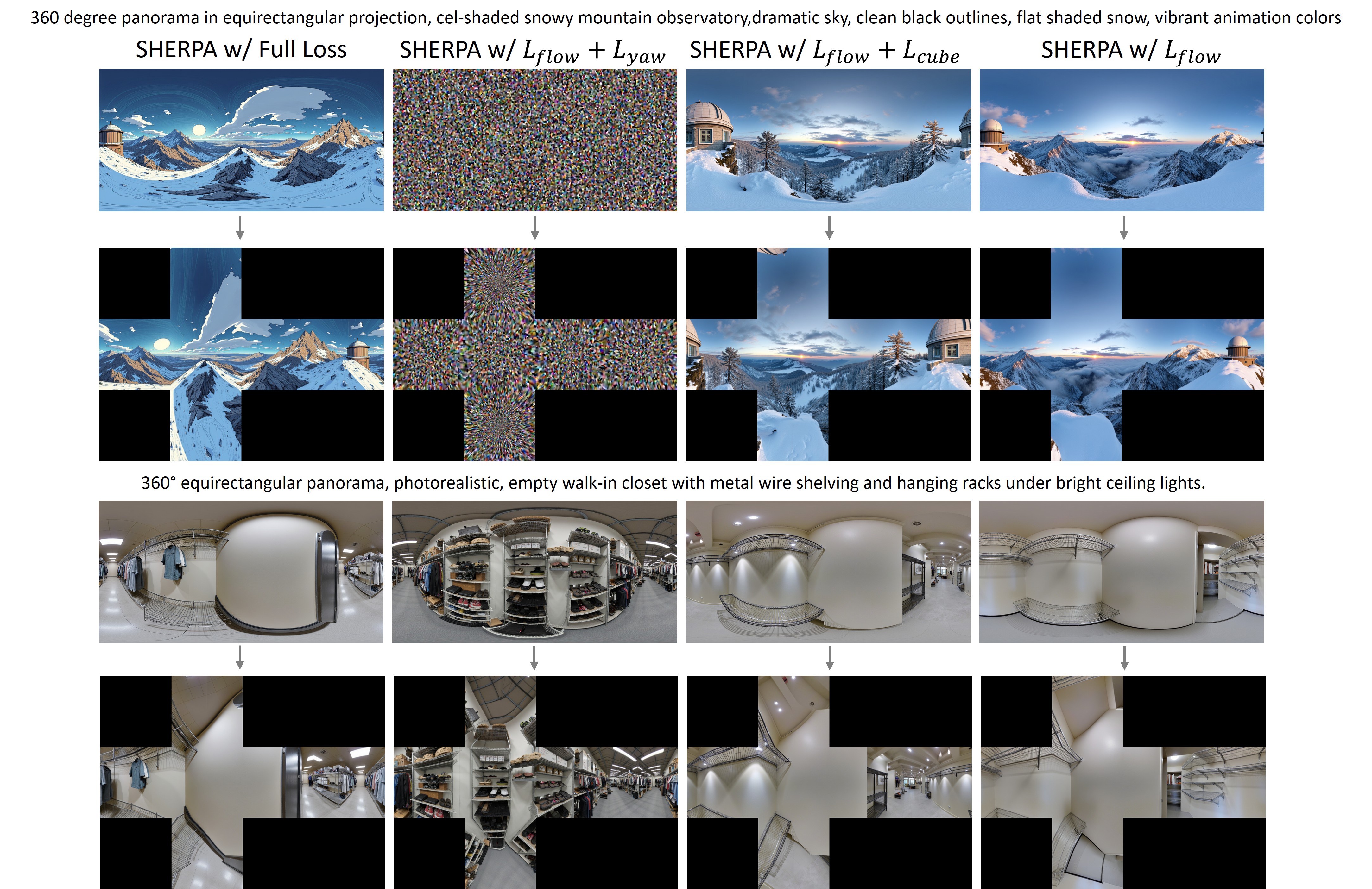}
\caption{\textbf{Loss ablation on stylized prompts.}
The full objective better preserves stylized panorama generation than geometry-only or yaw-only variants.}
\label{fig:loss_ablation}
\end{figure}

Table~\ref{tab:component_ablation} and Figs.~\ref{fig:parts_ablation}--\ref{fig:loss_ablation} summarize the ablations behind our design choices.
Circular RoPE targets seam topology, FFN adapters provide trainable panorama capacity, cubemap supervision anchors paired geometry, and yaw consistency addresses target-free style prompts.
The loss ablation shows the trade-off: cube-only training scores well on paired photorealistic metrics, whereas yaw without sufficient cubemap supervision or with excessive weight degrades image quality.

\paragraph{Structural seam closure.}
Figure~\ref{fig:parts_ablation} and Table~\ref{tab:component_ablation} isolate the non-trainable seam components.
Circular RoPE reduces seam MAE from 0.145 to 0.068 without adapter training, showing that much of the boundary mismatch is positional.
Circular Latent Encoding/Decoding is weaker alone but complements Circular RoPE as a decode-time boundary safeguard.

\paragraph{FFN adapter effect.}
RoPE+Latent E/D improves seam diagnostics but leaves FID high at 43.84.
Adding image-side FFN adapters with the flow objective reduces FID to 30.42, confirming that panorama-domain adaptation requires residual image-token transformations beyond attention-topology changes.
With cubemap-projected supervision, the FFN adapter also reaches the lowest polar FID in Table~\ref{tab:component_ablation}.

\paragraph{Cubemap-projected velocity supervision.}
The cubemap term re-expresses predicted and target velocities in local perspective faces, anchoring the adapter to valid panorama content.
This is crucial for target-free style prompts: flow+yaw without cube supervision degrades FID to 107.87 and worsens seam and polar metrics.
Cube-only training gives strong distributional metrics, but stylized prompts still need yaw consistency due to photorealistic data bias.

\paragraph{Unpaired Style Path yaw consistency.}
Yaw consistency discourages absolute azimuth bias for unpaired style prompts via cyclic-roll velocity consistency.
It is useful but delicate: an excessively strong yaw raises FID to 92.57 and geometry errors, whereas SHERPA improves stylized panoramas while staying close to cube-only FID.
This exposes a limitation of distributional metrics on paired photorealistic panorama data: cube-only training scores best on FID and polar FID, but Fig.~\ref{fig:loss_ablation} shows that style prompts are best preserved only when the full loss is used.
SHERPA therefore trades a small amount of paired-data distributional score for better target-free stylized panorama fidelity.
Together, Fig.~\ref{fig:loss_ablation} and Table~\ref{tab:component_ablation} show that yaw must be balanced by flow and cubemap supervision.

\section{Conclusion}

SHERPA shows that a strong planar text-to-image prior can be adapted into an open-domain panorama generator without surrendering stylized visual behavior.
By combining topology-aware structural changes, lightweight residual adaptation, and target-free yaw consistency, SHERPA balances panorama geometry with the creative range needed for world-building and simulation. Our analysis and ablations show that seam topology, residual panorama capacity, and unpaired style-prompt consistency play complementary roles rather than being interchangeable fixes.

\clearpage

\appendix
\section*{Appendix}

\section{Additional Analysis of Circular RoPE}
\label{app:circular_rope_analysis}

\subsection{Frequency-Band Analysis of FLUX RoPE}
\label{app:rope_band_analysis}

Circular RoPE is motivated by the observation that RoPE frequency bands can play different roles.
Recent analyses of RoPE suggest that high-frequency components are more closely associated with positional attention behavior, while lower-frequency components can contribute to broader semantic or global information~\cite{barbero2024round}.
Since those observations are not specific to FLUX or panorama generation, we analyze the width-axis RoPE behavior of FLUX in our setting.

The purpose of this analysis is to decide which part of the horizontal RoPE spectrum should be modified.
If all width-axis pairs are circularized, the pretrained positional spectrum can be excessively distorted.
If too few pairs are circularized, the panorama seam remains under-corrected.
We therefore separate the width-axis RoPE pairs into high-frequency and lower-frequency bands and use the high-frequency band as the candidate region for Circular RoPE.

\begin{figure}[p]
\centering
\includegraphics[width=0.85\textwidth]{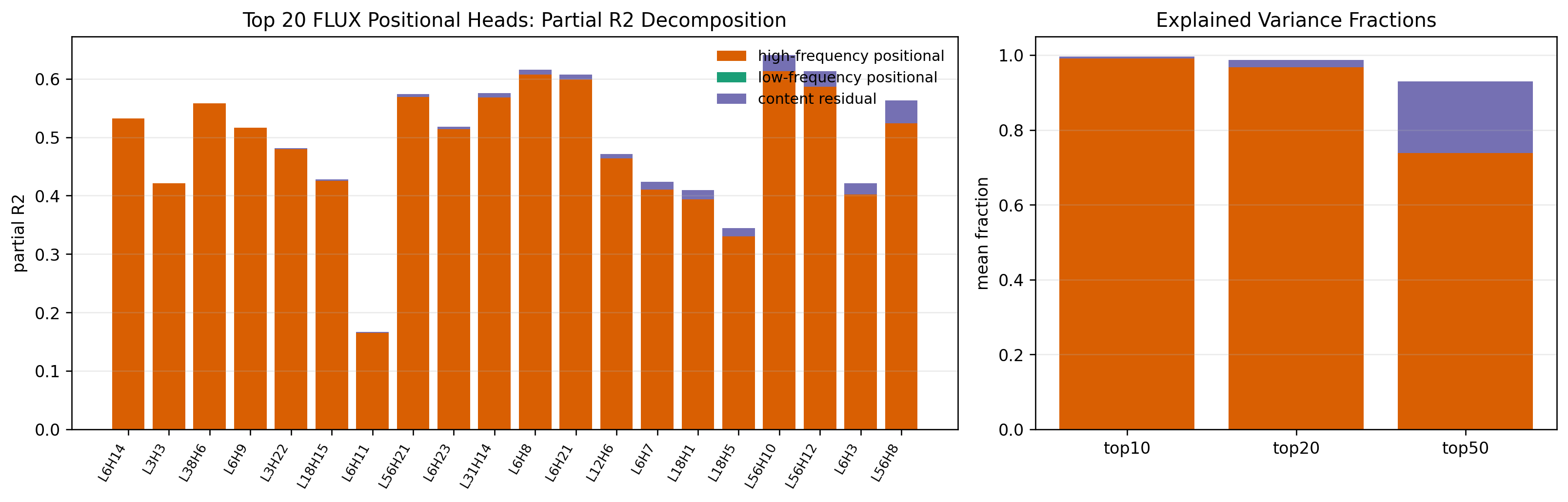}
\caption{\textbf{Frequency-band analysis of FLUX RoPE.}
Width-axis positional attribution is concentrated in high-frequency RoPE bands.}
\label{fig:app_rope_band_analysis}
\end{figure}

\subsection{Circular RoPE Phase Closure}
\label{app:phase_closure}

After identifying the high-frequency width-axis band as the target region, we circularize selected frequencies by replacing them with integer-periodic harmonics.
For each selected pair, shifting by one full panorama width changes the phase by an integer multiple of \(2\pi\):
\begin{equation}
\tilde{\omega}_j(c+W)
=
\tilde{\omega}_j c
+
2\pi m_j .
\end{equation}
Therefore, the selected RoPE pair is phase-closed across the panorama seam.

\begin{figure}[p]
\centering
\includegraphics[width=0.85\textwidth]{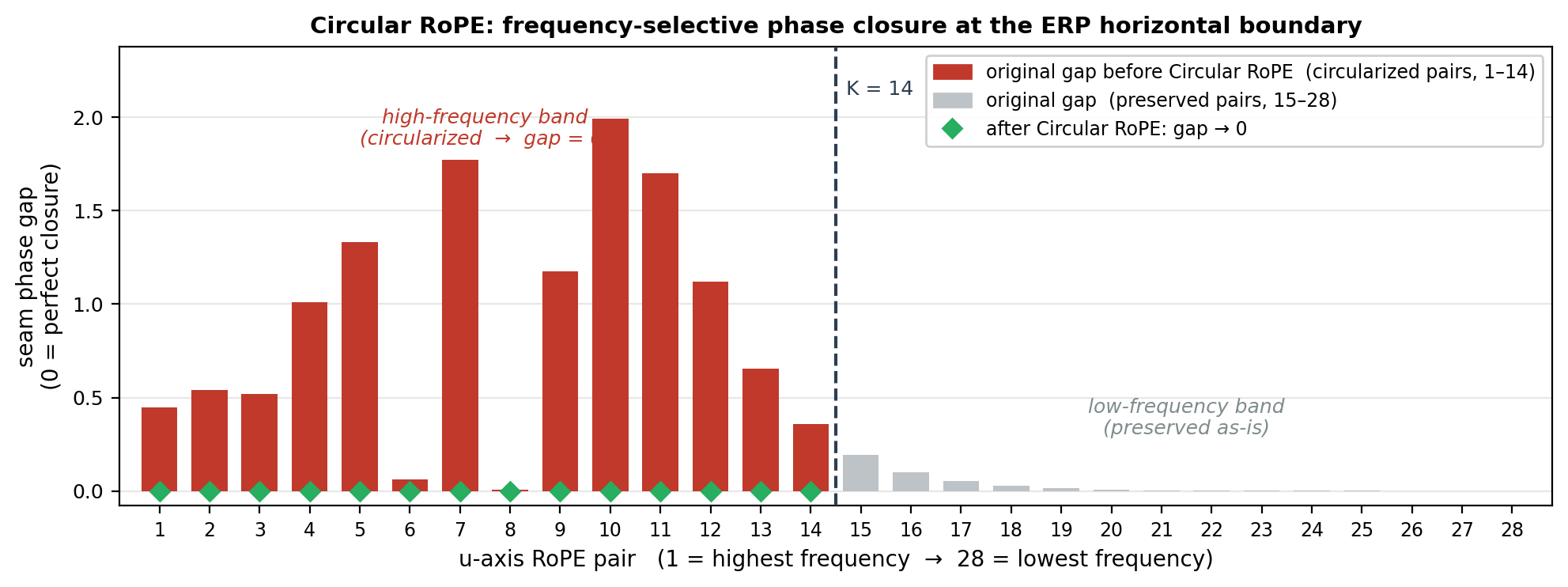}
\caption{\textbf{RoPE phase-closure diagnostic.}
Integer-periodic harmonics close the horizontal RoPE phase after one panorama width.}
\label{fig:app_phase_closure}
\end{figure}

\subsection{Circular RoPE Scope Sweep}
\label{app:rope_scope_sweep}

We sweep the number of circularized width-axis RoPE pairs in a RoPE-only diagnostic setting.
This sweep is not intended to evaluate the full SHERPA model.
Instead, it identifies the proper scope of the Circular RoPE rewrite before adding Circular Latent Encoding/Decoding, FFN adapters, and Unpaired Style Path regularization.

Let \(K\) denote the number of circularized high-frequency width-axis RoPE pairs.
Small \(K\) values minimally perturb the pretrained RoPE spectrum, but they can under-correct the panorama seam.
Larger \(K\) values provide stronger seam correction, but once the rewrite enters the lower-frequency band, it can corrupt the pretrained global visual behavior.
In FLUX, the width-axis RoPE contains 28 pairs.
The setting \(K=14\) corresponds to circularizing the complete high-frequency half while preserving the lower-frequency half.
We therefore use \(K=14\) as the default Circular RoPE setting.

\begin{figure}[p]
\centering
\includegraphics[width=0.9\textwidth]{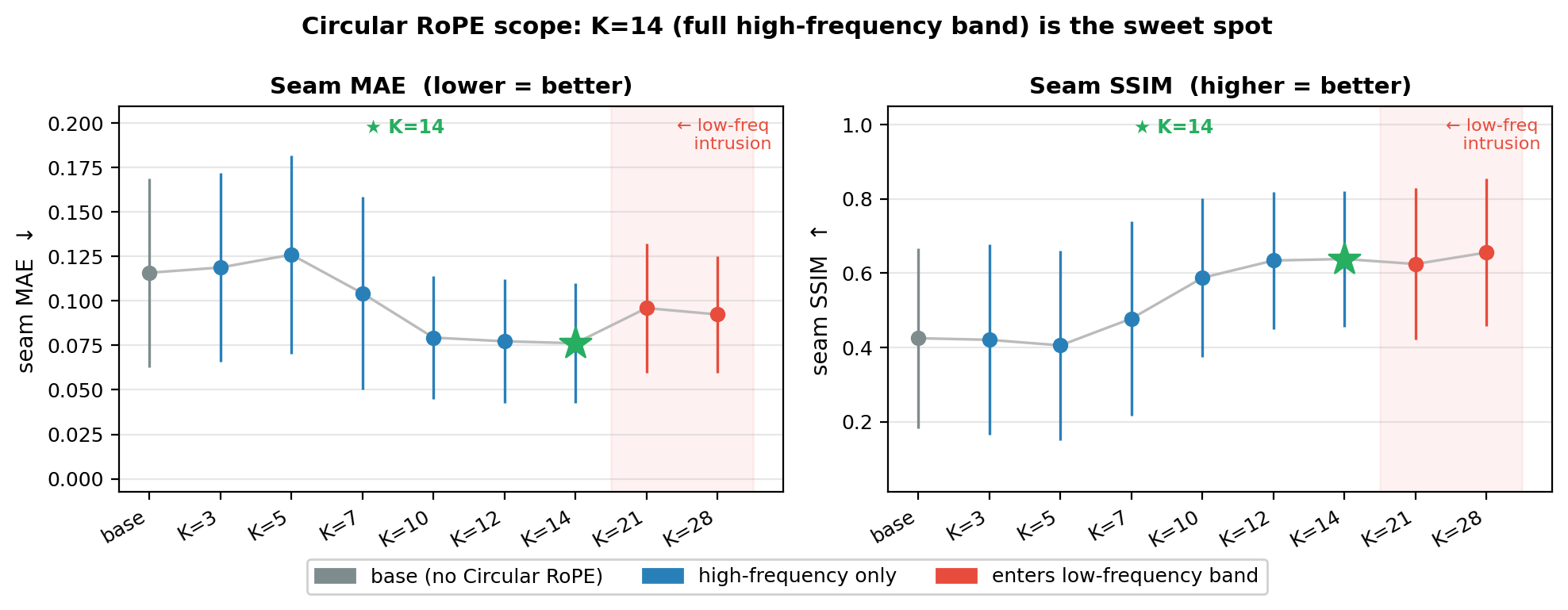}
\caption{\textbf{Circular RoPE frequency-scope sweep.}
We sweep the number of circularized width-axis RoPE pairs and use \(K=14\) by default.}
\label{fig:app_rope_k_sweep}
\end{figure}

\subsection{Why Horizontal RoPE Correction Does Not Solve Pole Distortion}
\label{app:seam_pole}

Circular RoPE is designed to correct the horizontal seam topology of panoramas.
Here we explain why we do not attempt to solve both seam closure and pole collapse by modifying RoPE alone.

\paragraph{RoPE phase and attention.}
For a rotary pair \(d\), RoPE applies a phase rotation to the query and key channels.
We write the phase at panorama token coordinate \((u,v)\) as
\begin{equation}
\phi_d(u,v),
\end{equation}
where \(u\in\{0,\ldots,W-1\}\) is the horizontal coordinate and \(v\in\{0,\ldots,H-1\}\) is the vertical coordinate.
The attention logit between two tokens depends on their relative phase:
\begin{equation}
\tilde{q}_{u_1,v_1}^{\top}\tilde{k}_{u_2,v_2}
=
q_{u_1,v_1}^{\top}
R\!\left(\phi_d(u_2,v_2)-\phi_d(u_1,v_1)\right)
k_{u_2,v_2}.
\end{equation}
Thus, if two panorama coordinates should represent the same point on the viewing sphere, their phase difference should be an integer multiple of \(2\pi\).

\paragraph{Seam condition.}
Panoramas are horizontally periodic.
The left and right boundaries correspond to adjacent longitudes, so the phase should close after one horizontal period:
\begin{equation}
\phi_d(u+W,v)-\phi_d(u,v)
\equiv 0
\pmod{2\pi}
\qquad
\forall u,v .
\label{eq:app_seam_condition}
\end{equation}
For a RoPE-like phase that is linear in the horizontal coordinate,
\begin{equation}
\phi_d(u,v)=f_d(v)u+g_d(v),
\label{eq:app_latitude_freq}
\end{equation}
the seam condition becomes
\begin{equation}
f_d(v)W
\equiv 0
\pmod{2\pi}
\qquad
\forall v .
\end{equation}
Equivalently,
\begin{equation}
f_d(v)
=
\frac{2\pi n_d(v)}{W},
\qquad
n_d(v)\in\mathbb{Z}.
\label{eq:app_seam_lattice}
\end{equation}
Thus, exact seam closure requires the horizontal frequency at each latitude to lie on an integer-periodic harmonic lattice.

\paragraph{Pole condition.}
At the north and south poles, all longitudes collapse to a single point on the sphere.
Therefore, for the top row \(v=0\), all horizontal positions should have identical phase:
\begin{equation}
\phi_d(u_1,0)-\phi_d(u_2,0)
\equiv 0
\pmod{2\pi}
\qquad
\forall u_1,u_2 .
\label{eq:app_pole_condition}
\end{equation}
Using Eq.~\eqref{eq:app_latitude_freq}, this requires
\begin{equation}
f_d(0)(u_1-u_2)
\equiv 0
\pmod{2\pi}
\qquad
\forall u_1,u_2 .
\end{equation}
In the principal, non-aliased representation, this corresponds to
\begin{equation}
f_d(0)=0.
\label{eq:app_pole_zero_north}
\end{equation}
The same condition holds at the south pole:
\begin{equation}
f_d(H-1)=0.
\label{eq:app_pole_zero_south}
\end{equation}

\paragraph{Conflict under smooth RoPE-like rewrites.}
Equations~\eqref{eq:app_seam_lattice}, \eqref{eq:app_pole_zero_north}, and \eqref{eq:app_pole_zero_south} reveal the core conflict.
Exact seam closure requires \(f_d(v)\) to take values in the discrete lattice
\begin{equation}
\frac{2\pi}{W}\mathbb{Z},
\end{equation}
while pole collapse requires the horizontal frequency to vanish at the top and bottom rows.
If \(f_d(v)\) is continuous in \(v\), then the integer-valued function \(n_d(v)\) in Eq.~\eqref{eq:app_seam_lattice} must be constant over \(v\).
Because the pole condition requires \(n_d(0)=0\) and \(n_d(H-1)=0\), continuity implies
\begin{equation}
n_d(v)\equiv 0
\qquad
\Longrightarrow
\qquad
f_d(v)\equiv 0 .
\end{equation}
This is a degenerate solution: the horizontal RoPE phase carries no non-trivial positional variation.

A non-degenerate solution would require \(n_d(v)\) to change across latitude.
However, since \(n_d(v)\) is integer-valued, such a change must occur through discontinuous jumps in the horizontal frequency.
These jumps break the smooth positional structure expected by the pretrained model and can introduce vertical inconsistency.
Therefore, a smooth RoPE-like frequency rewrite cannot simultaneously provide exact horizontal seam closure and exact pole collapse in a non-degenerate way.

\paragraph{Latitude-scaled RoPE.}
A natural alternative is to scale the horizontal frequency by latitude:
\begin{equation}
\phi_d^{\mathrm{lat}}(u,v)
=
k_d\frac{2\pi}{W}\sin\theta(v)\,u,
\qquad
\theta(v)=\pi\frac{v}{H-1}.
\label{eq:app_lat_scaled}
\end{equation}
This satisfies the pole condition because
\begin{equation}
\sin\theta(0)=\sin\theta(H-1)=0.
\end{equation}
However, it generally violates the seam condition.
Exact seam closure requires
\begin{equation}
k_d\sin\theta(v)\in\mathbb{Z}
\qquad
\forall v .
\label{eq:app_lat_scaled_seam}
\end{equation}
This condition does not hold for general latitudes.
For example, at \(\theta=\pi/4\),
\begin{equation}
\sin\theta=\frac{\sqrt{2}}{2},
\end{equation}
so \(k_d\sin\theta\) is generally not an integer for integer \(k_d\).
Thus, latitude scaling can make pole behavior more sphere-like, but it breaks exact seam phase closure at most latitudes.

\paragraph{Implication.}
This analysis motivates the scope of Circular RoPE.
We use RoPE to solve the part of panorama geometry that is naturally a phase-closure problem: the horizontal seam.
We do not attempt to solve pole collapse by circularizing the vertical axis or applying a full spherical RoPE rewrite.
Instead, residual pole and global-structure issues are left to the learned FFN adapters.

\begin{figure}[t]
\centering
\includegraphics[width=\textwidth]{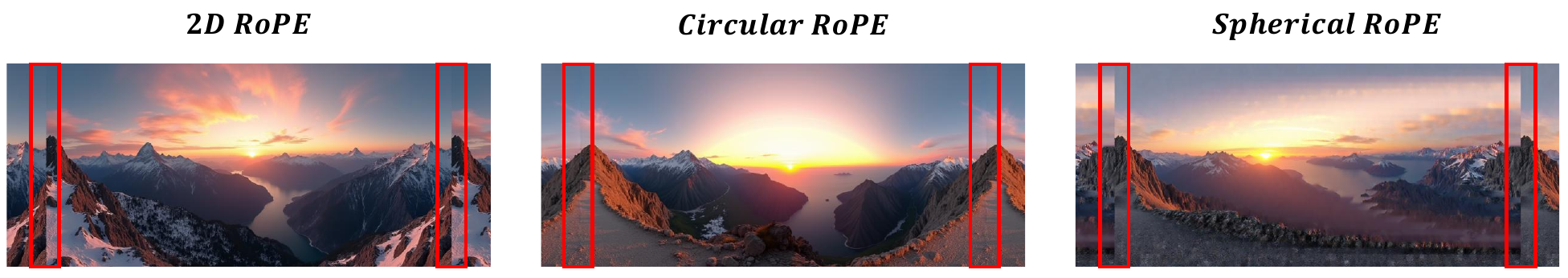}
\caption{\textbf{Horizontal versus spherical RoPE rewrite.}
Circular RoPE modifies the horizontal axis, while spherical rewrites modify both horizontal and vertical axes.}
\label{fig:app_rope_compare}
\end{figure}

\section{Implementation and Asset Details}
\label{app:compute_assets}

\paragraph{Compute resources.}
All SHERPA adapter experiments were run on local NVIDIA RTX PRO 5000 Blackwell GPUs with 48GB memory.
The final adapter uses rank-32 image-stream FFN adapters in the 38 FLUX single-stream blocks, for 7,471,104 trainable parameters, corresponding to about \(0.062\%\) of the 12B-parameter FLUX.1-dev backbone; the FLUX backbone is frozen.
The reported experiments include the final adapter training and the ablations needed to support the design choices in the main text.
We keep the pretrained FLUX backbone frozen and optimize only the adapter parameters.
From training logs and checkpoint timestamps, the reported SHERPA checkpoint required approximately 7.4 GPU-hours on one RTX PRO 5000 Blackwell GPU, including a 500-step warm start and continuation to the selected 2,500-step checkpoint.
The same run was later continued to 4,000 steps for analysis, taking approximately 10.9 GPU-hours in total, but the 2,500-step checkpoint is the one used for the main results.

\begin{table}[h]
\centering
\scriptsize
\renewcommand{\arraystretch}{1.08}
\begin{tabularx}{\textwidth}{@{}lX@{}}
\toprule
\textbf{Item} & \textbf{Setting}\\
\midrule
Backbone and trainable modules & Frozen FLUX.1-dev backbone; rank-32 image-stream FFN adapters in all 38 single-stream blocks.\\
Training resolution & \(1024{\times}512\) panorama images, encoded to \(64{\times}32\) latent grids before packing.\\
Optimizer and learning rate & AdamW, learning rate \(1{\times}10^{-4}\).\\
Training length & 2,500 optimizer steps for the reported SHERPA checkpoint; ablation variants use the steps reported in Table~\ref{tab:component_ablation}.\\
Paired Panorama Path & One paired panorama batch per step from the polished Matterport3D pool of 10,359 images, with rectified-flow loss, cubemap-projected velocity loss, and horizontal shift augmentation.\\
Unpaired Style Path & One unpaired style prompt batch per step from a 108-prompt style pool; yaw consistency uses original and horizontally rolled random latents.\\
Loss weights & Reported SHERPA uses polar/equatorial cubemap weights \(0.5\), yaw weight decayed from \(75\) to \(0\), and no base-prior regularization.\\
Batch ratio & \(1{:}1\) paired panorama to unpaired style batches per optimizer step.\\
Inference & 25 sampling steps and guidance scale 3.5 unless a baseline pipeline fixes different defaults.\\
Evaluation split & 517 held-out photorealistic paired Matterport test prompts/images per seed for Table~\ref{tab:main_quant}; the user study uses five prompt/style settings covering four stylized prompts and one photorealistic prompt.\\
\bottomrule
\end{tabularx}
\caption{\textbf{Key implementation settings for reproducibility.}
The ablation table varies only the listed loss weights or structural components unless otherwise noted.}
\label{tab:app_hparams}
\end{table}

\begin{table}[h]
\centering
\scriptsize
\renewcommand{\arraystretch}{1.08}
\begin{tabular}{@{}lcccc@{}}
\toprule
\textbf{Model} & \textbf{Resolution} & \textbf{Steps} & \textbf{Guidance} & \textbf{Time/image}\\
\midrule
Base FLUX & \(1024{\times}512\) & 25 & 3.5 & \(6.758 \pm 0.054\) s\\
SHERPA & \(1024{\times}512\) & 25 & 3.5 & \(6.849 \pm 0.005\) s\\
\bottomrule
\end{tabular}
\caption{\textbf{Inference speed measurement.}
Measured on one RTX PRO 5000 Blackwell GPU after one warm-up image, excluding model loading.
SHERPA adds 1.34\% latency over base FLUX.}
\label{tab:app_inference_speed}
\end{table}

\paragraph{Inference and evaluation protocol.}
We generate or collect baseline outputs from official checkpoints/repositories when available, using the shared prompt list and fixed seeds whenever the pipeline exposes seed control.
If a method has a different native output size, we resize only for metric computation and apply the same panorama-to-cubemap preprocessing to all methods.
FID and IS are computed on cubemap faces; polar FID uses the top and bottom cubemap faces; CLIP is computed after resizing the whole panorama as a single image.
The same evaluation scripts and preprocessing are used for SHERPA, ablations, and baselines.

\paragraph{Existing assets and licenses.}
We use FLUX.1-dev as the pretrained backbone~\cite{flux2024}; the official Black Forest Labs repository lists FLUX.1-dev under the FLUX.1-dev Non-Commercial License, while the associated autoencoder weights are listed as Apache-2.0.
For paired panorama supervision, we use the polished Matterport3D version/split adopted by DiT360~\cite{feng2025dit360}, which is derived from Matterport3D~\cite{chang2017matterport3d}.
We follow the Matterport academic and non-commercial research terms for the underlying Matterport3D data.
Poly Haven prompts and references are derived from Poly Haven assets, whose official asset license is CC0~\cite{polyhaven}.
SUN360~\cite{xiao2012sun360} and Laval HDR panoramas~\cite{gardner2017learning} are cited as representative panorama datasets used in prior work, not as additional SHERPA training data.
The SUN360 project page states that its images and materials are for academic research use only, and the Laval HDR datasets are distributed under custom research/data-use terms.
We therefore treat them as restricted research datasets and do not redistribute their images or derived copies.
CLIP-based metrics use OpenAI CLIP~\cite{radford2021clip}, whose official code repository is released under the MIT License.
Baseline methods are credited to their original papers and repositories; any released code, checkpoints, or derived assets will be distributed only when permitted by the corresponding upstream licenses and access terms.

\paragraph{Broader impacts.}
Panoramic generation is increasingly relevant to world models, VR/AR, simulation, games, robotics, and spatial content creation.
By combining a lightweight adapter with a pretrained FLUX backbone, SHERPA may lower the cost of adapting open-domain image priors to diverse panorama scenes without training a generator from scratch.
It also inherits general text-to-image risks, including misleading synthetic scenes, immersive-media misuse, and biases from the pretrained backbone and panorama data; generated panoramas should be disclosed as synthetic in public-facing uses.

\paragraph{Limitations.}
SHERPA remains fixed to a \(1024{\times}512\) ERP latent grid and does not fully remove pole distortion.
Broader DiT architectures, higher resolutions, and native spherical or 3D representations remain future work.

\section{Additional Qualitative Comparisons}
\label{app:additional_qual}
\begin{figure}[p]
\centering
\includegraphics[width=\textwidth]{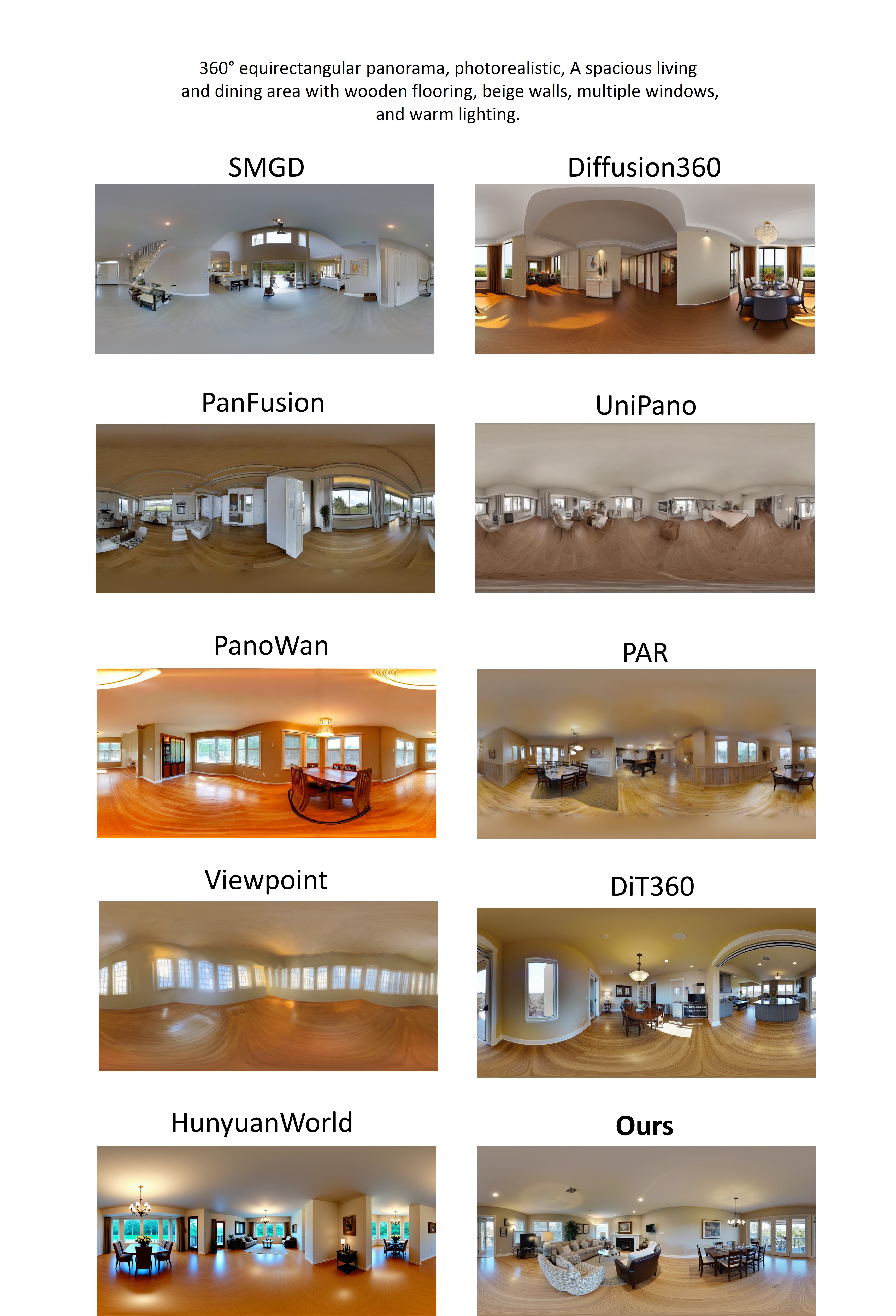}
\caption{\textbf{Additional open-domain comparison.}
Additional comparison with panorama generation baselines.}
\label{fig:compare1}
\end{figure}

\begin{figure}[p]
\centering
\includegraphics[width=\textwidth]{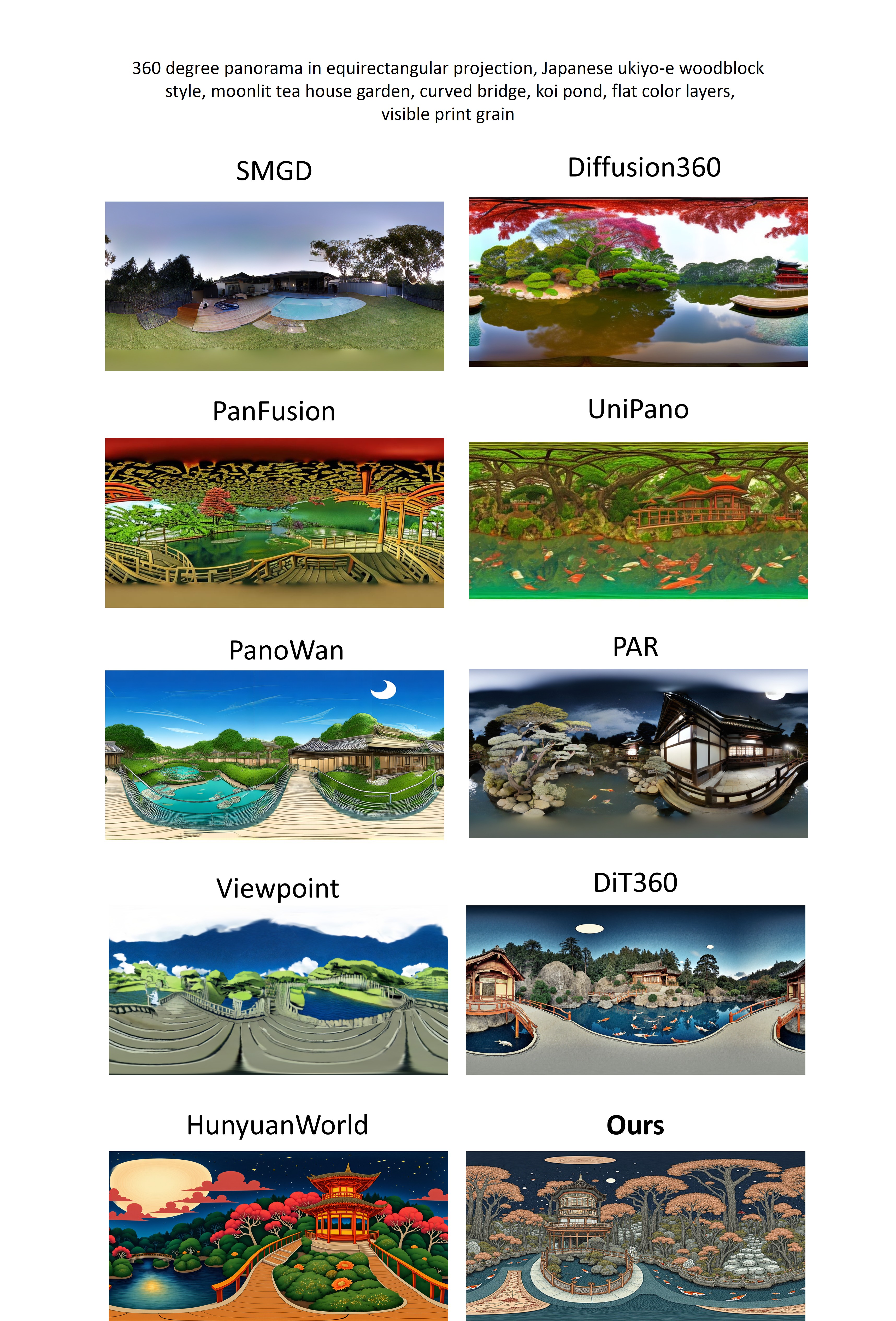}
\caption{\textbf{Additional open-domain comparison.}
Additional comparison with panorama generation baselines.}
\label{fig:compare12}
\end{figure}

Figures~\ref{fig:compare1}--\ref{fig:compare12} illustrate the performance of SHERPA relative to the baselines across a diverse range of text styles.

\begin{figure}[p]
\centering
\includegraphics[width=\textwidth]{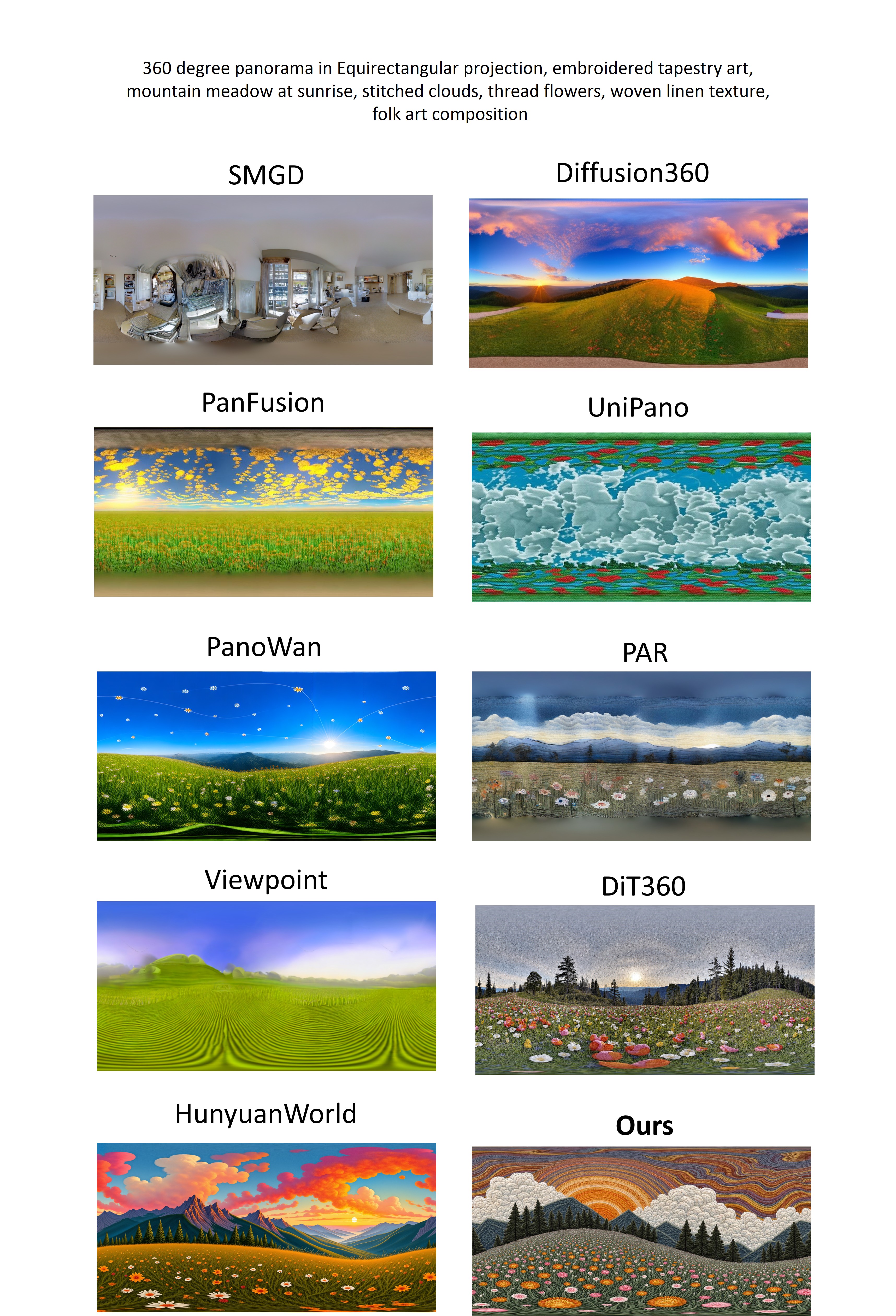}
\caption{\textbf{Additional open-domain comparison.}
Additional comparison with panorama generation baselines.}
\end{figure}

\begin{figure}[p]
\centering
\includegraphics[width=\textwidth]{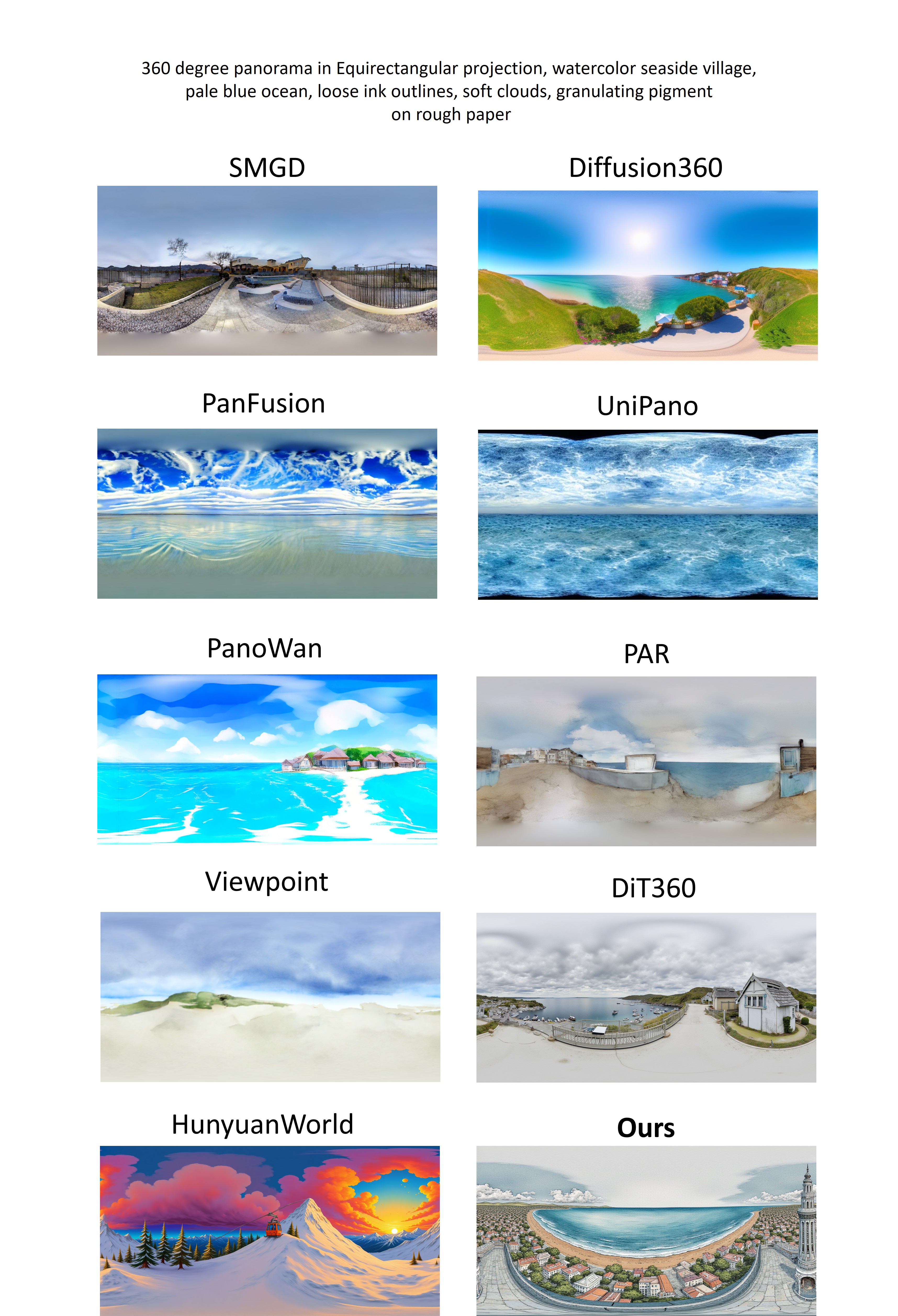}
\caption{\textbf{Additional open-domain comparison.}
Additional comparison with panorama generation baselines.}
\end{figure}

\begin{figure}[p]
\centering
\includegraphics[width=\textwidth]{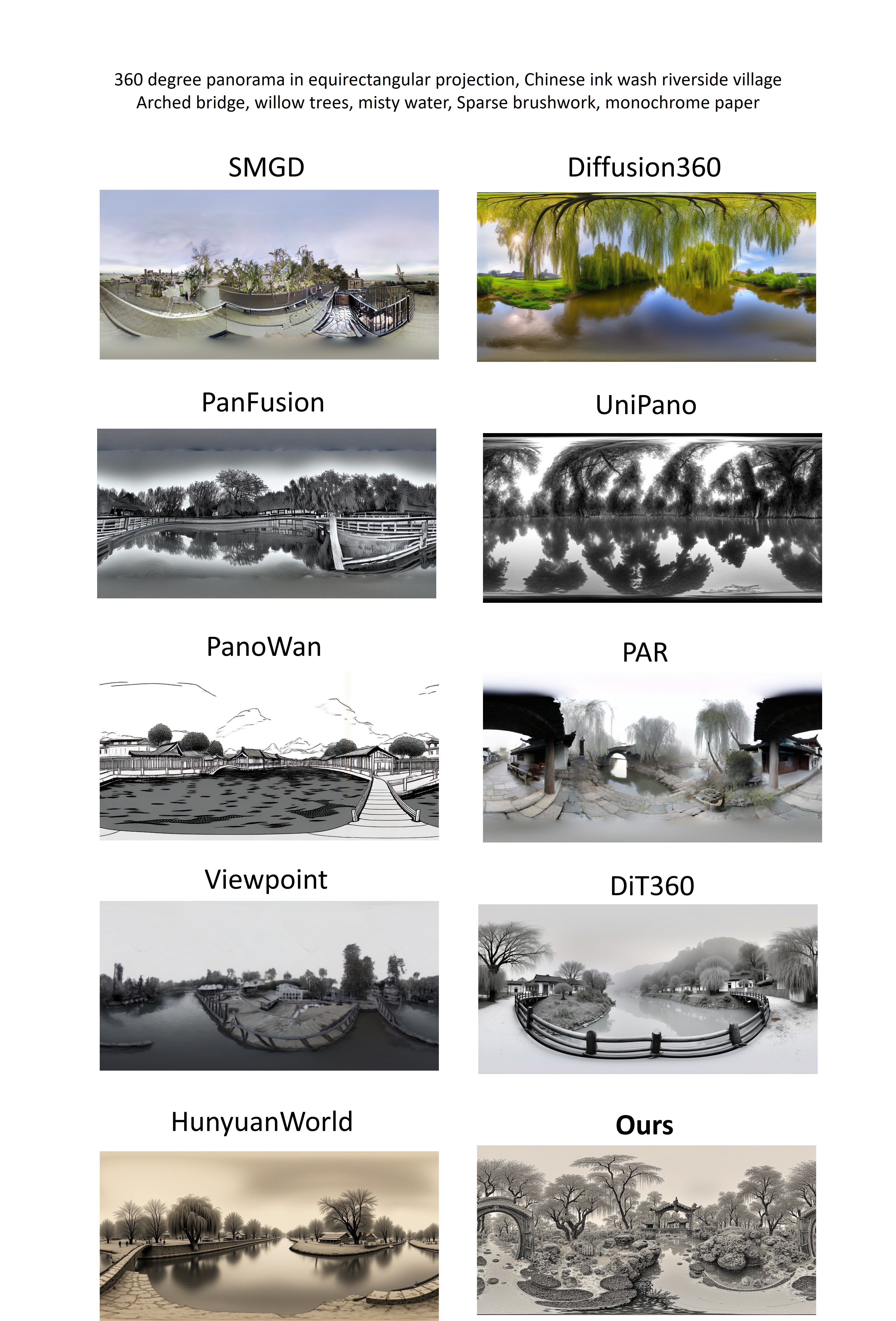}
\caption{\textbf{Additional open-domain comparison.}
Additional comparison with panorama generation baselines.}
\end{figure}

\begin{figure}[p]
\centering
\includegraphics[width=\textwidth]{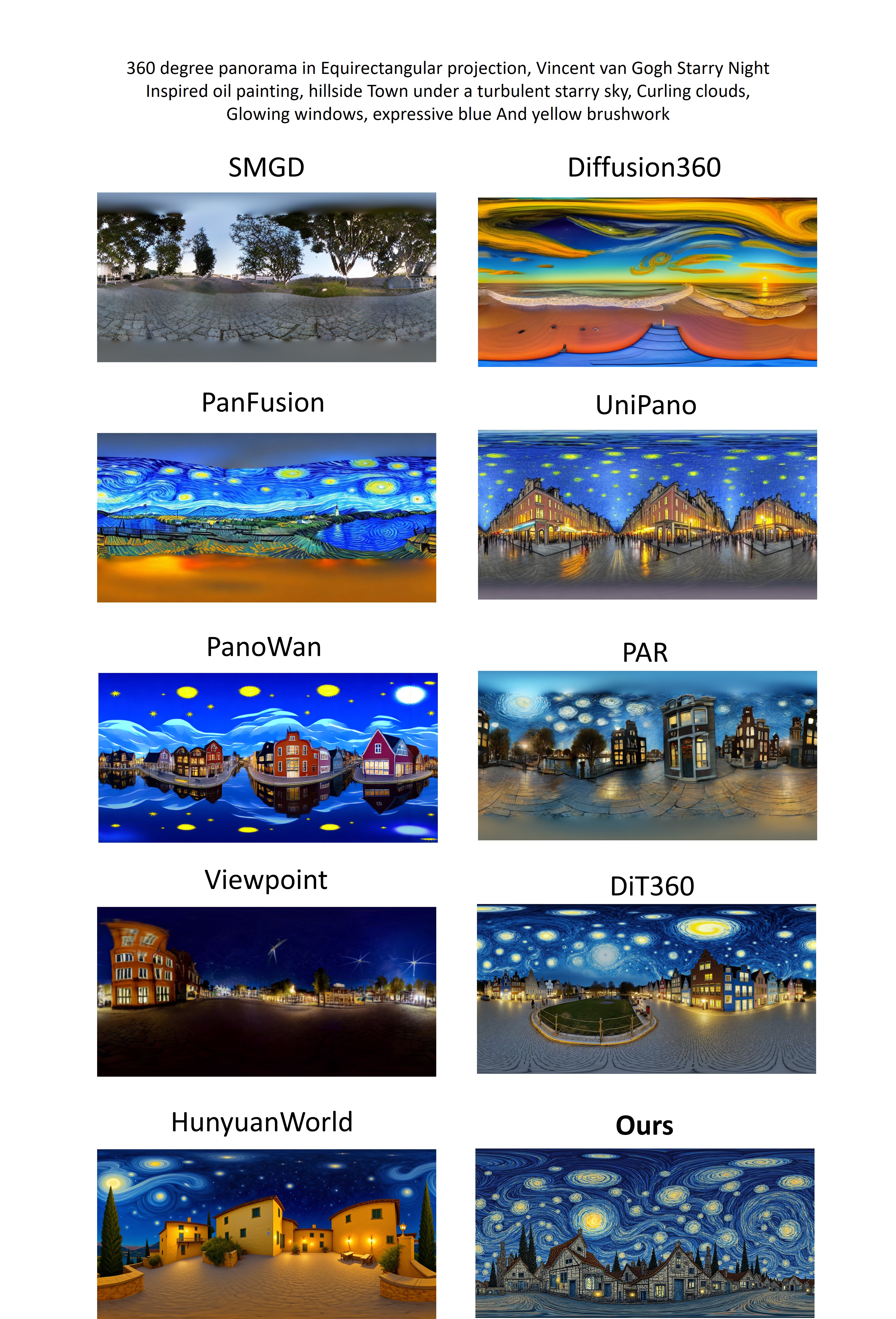}
\caption{\textbf{Additional open-domain comparison.}
Additional comparison with panorama generation baselines.}
\end{figure}

\begin{figure}[p]
\centering
\includegraphics[width=\textwidth]{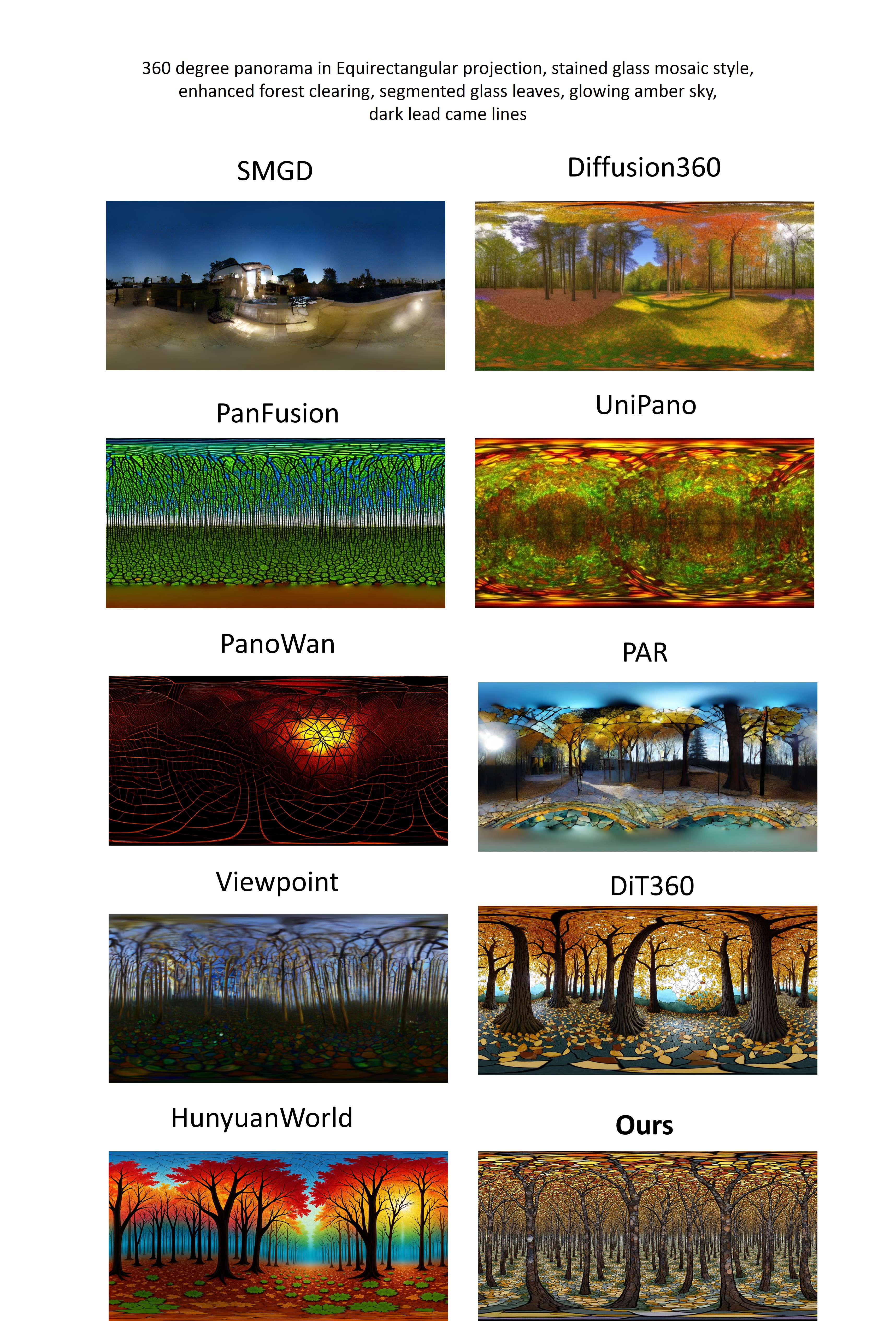}
\caption{\textbf{Additional open-domain comparison.}
Additional comparison with panorama generation baselines.}
\end{figure}

\begin{figure}[p]
\centering
\includegraphics[width=\textwidth]{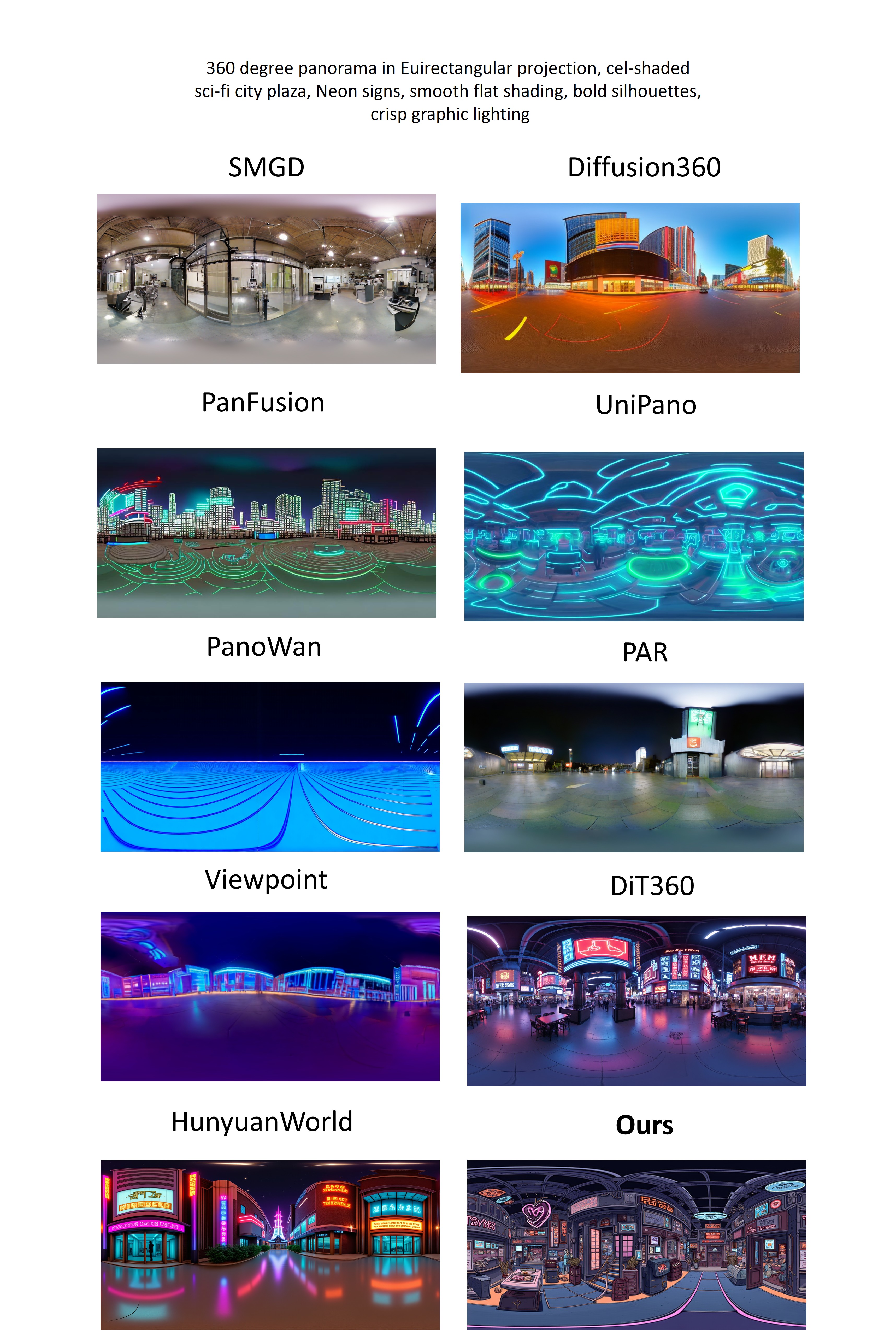}
\caption{\textbf{Additional open-domain comparison.}
Additional comparison with panorama generation baselines.}
\end{figure}

\begin{figure}[p]
\centering
\includegraphics[width=\textwidth]{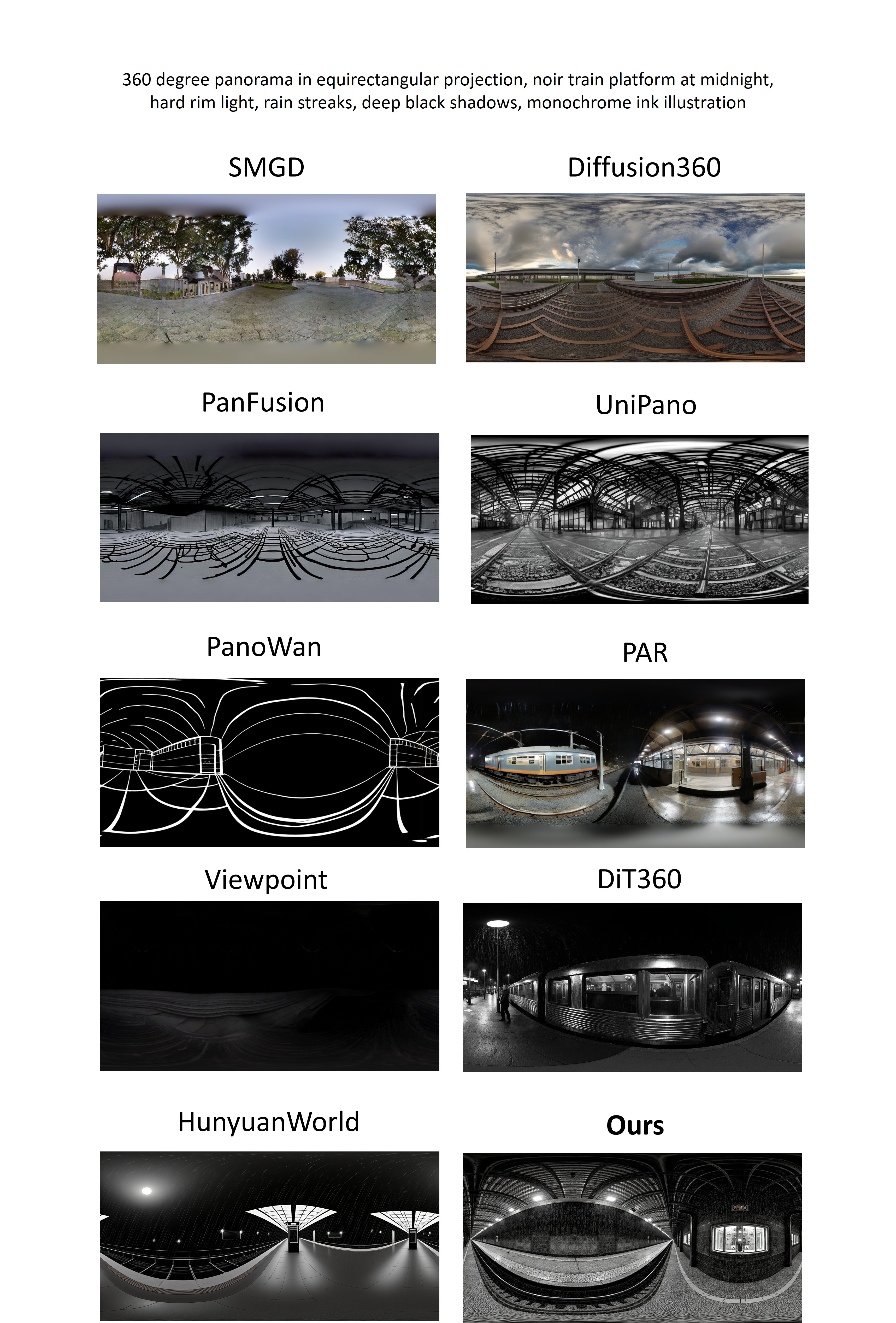}
\caption{\textbf{Additional open-domain comparison.}
Additional comparison with panorama generation baselines.}
\end{figure}

\begin{figure}[p]
\centering
\includegraphics[width=\textwidth]{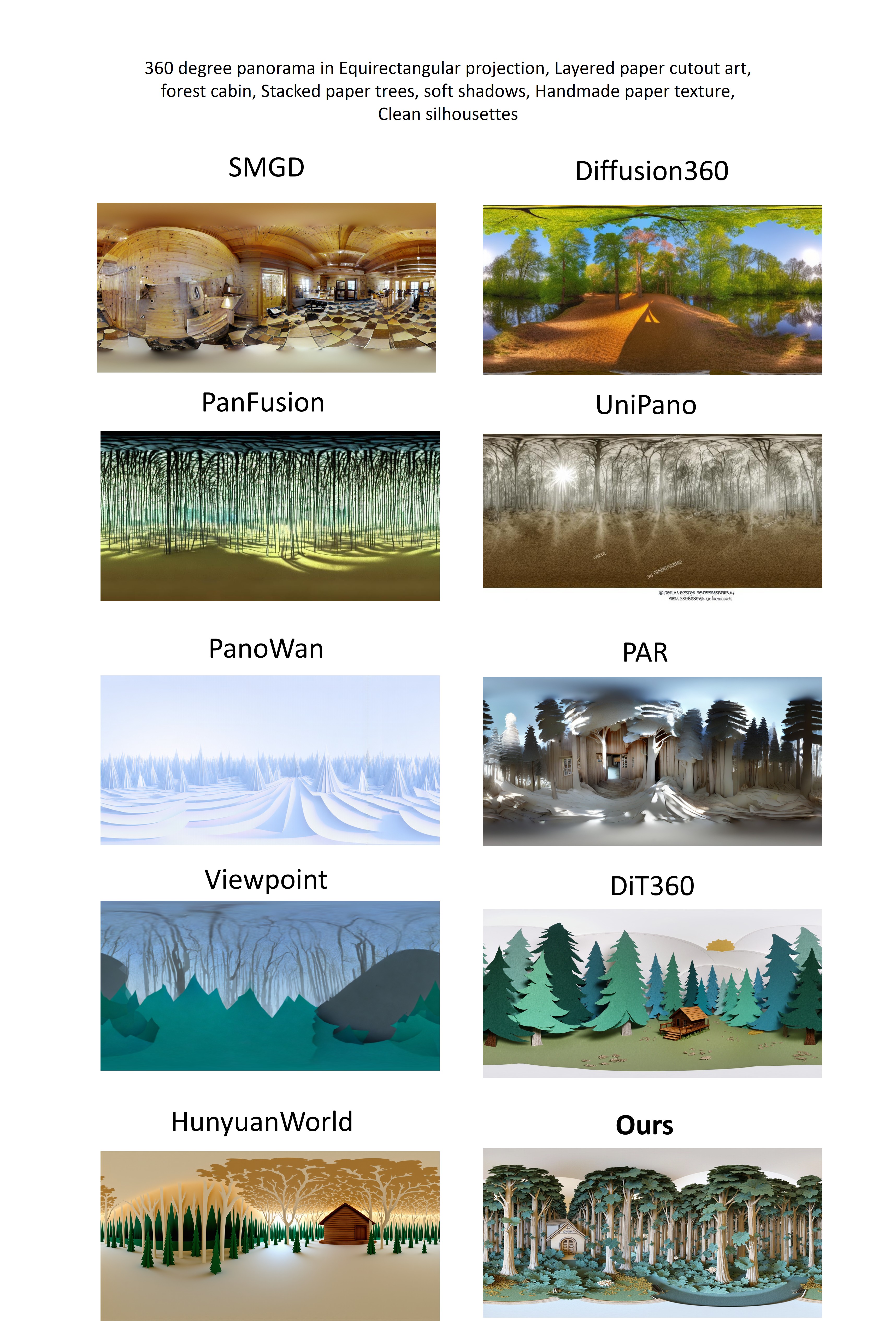}
\caption{\textbf{Additional open-domain comparison.}
Additional comparison with panorama generation baselines.}
\end{figure}

\begin{figure}[p]
\centering
\includegraphics[width=\textwidth]{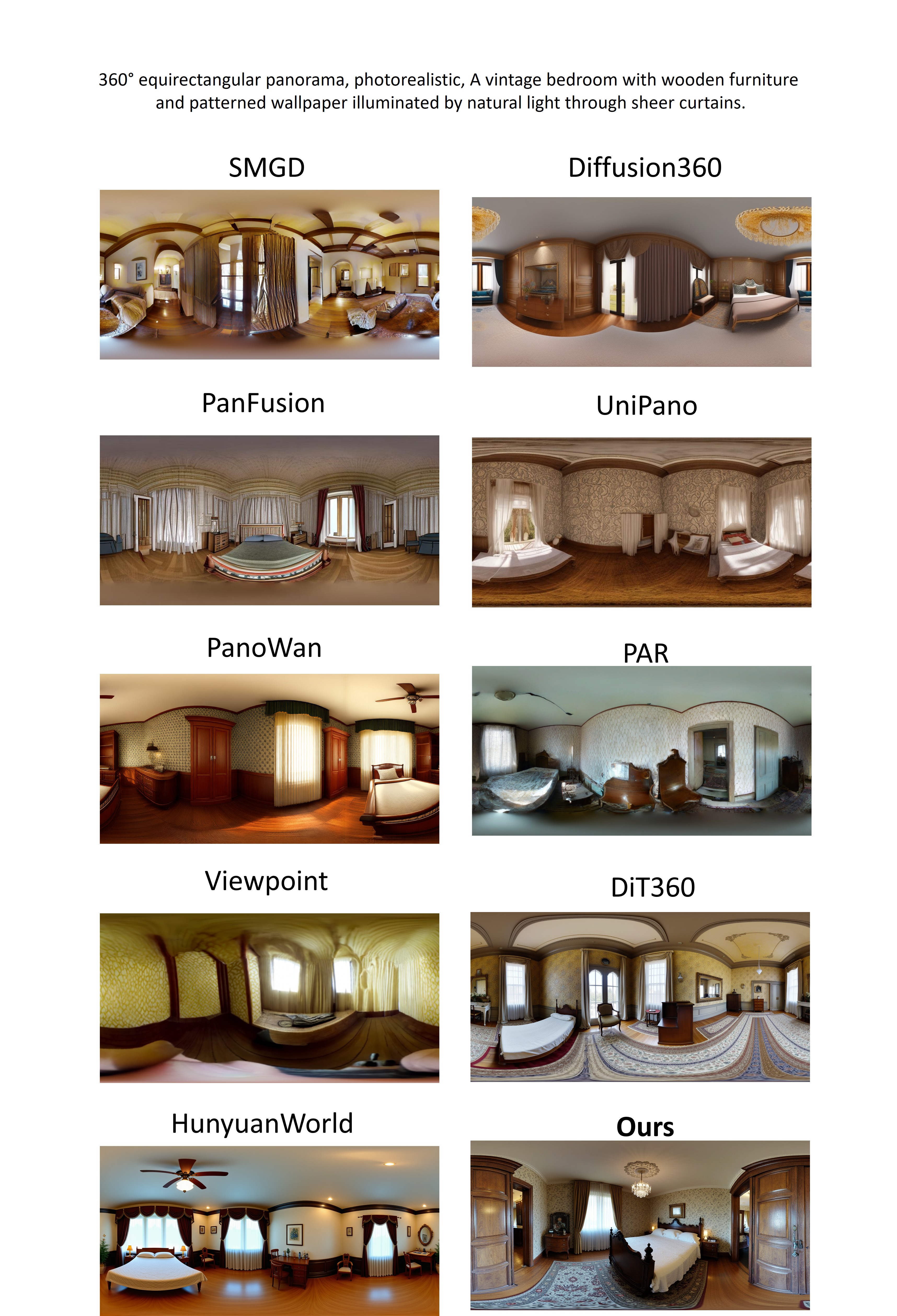}
\caption{\textbf{Additional open-domain comparison.}
Additional comparison with panorama generation baselines.}
\end{figure}

\begin{figure}[p]
\centering
\includegraphics[width=\textwidth]{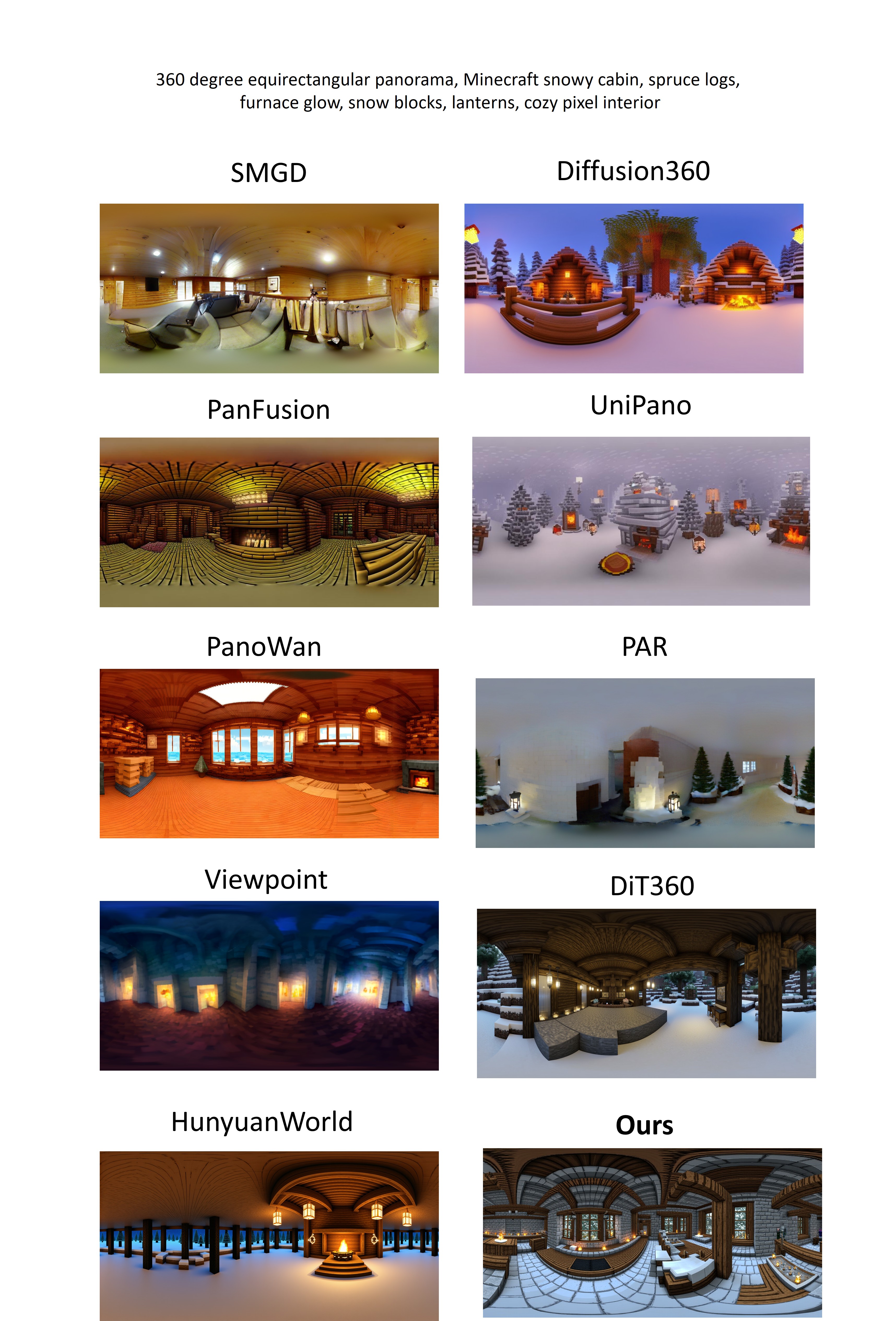}
\caption{\textbf{Additional open-domain comparison.}
Additional comparison with panorama generation baselines.}
\end{figure}

\clearpage

\end{document}